\documentclass[twocolumn]{IEEEtran}
\IEEEoverridecommandlockouts
\usepackage{textcomp,hyperref}       % hyperlinks
\usepackage{url}            % simple URL typesetting
\usepackage{booktabs}       % professional-quality tables
\usepackage{amsfonts}       % blackboard math symbols
\usepackage{nicefrac}       % compact symbols for 1/2, etc.
\usepackage{cite}
\usepackage{amsmath,amssymb,amsthm,epsfig}
\usepackage{algorithmic}
\usepackage{subcaption,graphicx}
\usepackage{bbm}
\usepackage{tikz, verbatim}
\usepackage{subfiles}
\usepackage[linesnumbered,ruled,vlined]{algorithm2e}
\usepackage{array}

\newcommand{\EXP}[1]{\mathsf{E}\!\left(#1\right)}

\newcommand{\DM}[1]{}
\newcommand{\SM}[1]{}
\newcommand{\VN}[1]{}
\newcommand{\remove}[1]{\textcolor{red}{removed content: commented
      out in source.}}

\newtheorem{thm}{Theorem}%[section]
\newtheorem{lem}{Lemma}

\newtheorem{defn}{Definition}%[section]

\title{Influencing Bandits: \\Arm Selection for Preference Shaping
  \thanks{The authors are also affiliated with the Bharti Centre in
    IIT Bombay. This work was supported in part by DST and CEFIPRA} }

\author{\IEEEauthorblockN{Viraj Nadkarni, D. Manjunath, and Sharayu Moharir}

  \IEEEauthorblockA{Department of of Electrical Engineering, IIT Bombay }
  
  nadkarniv4198@gmail.com, dmanju,sharayu@ee.iitb.ac.in
}

\begin{document}
\maketitle

\begin{abstract}
  We consider a non stationary multi-armed bandit in which the
  population preferences are positively and negatively reinforced by
  the observed rewards. The objective of the algorithm is to shape
  the population preferences to maximize the fraction of the
  population favouring a predetermined arm. For the case of binary
  opinions, two types of opinion dynamics are considered---decreasing
  elasticity (modeled as a Polya urn with increasing number of balls)
  and constant elasticity (using the voter model). For the first case,
  we describe an Explore-then-commit policy and a Thompson sampling
  policy and analyse the regret for each of these policies. We then
  show that these algorithms and their analyses carry over to the
  constant elasticity case. We also describe a Thompson sampling based
  algorithm for the case when more than two types of opinions are
  present. Finally, we discuss the case where presence of multiple
  recommendation systems gives rise to a trade-off between their
  popularity and opinion shaping objectives.
\end{abstract}
\textbf{Keywords:Multi-armed Bandits, Opinion Shaping, Contextual
  Bandits, Non-stationary rewards}

\section{Introduction}\label{sec:intro}
Stochastic multi-armed bandit (MAB) algorithms are used in many
applications. A canonical application is in recommendation systems
that suggest one or more items from a fixed set of items to users of
the system. Example uses of these recommendation systems are for
suggesting items on e-commerce websites and on streaming services,
recommending friends and connections on social networks, and for
placing results and ads in search engines. The classical stochastic
MAB setting is as follows. Users arrive sequentially and are
recommended one of $N$ arms. The system obtains a random reward based
on the user preference for the recommended arm. The reward
distribution of each arm is assumed independent of all other rewards
and is unknown. The objective of the MAB algorithm is to learn the
best arm while minimizing the loss in cumulative reward, compared to
that from an ideal or a reference algorithm, over a finite time
horizon. Among the many extensions to this classical problem, the one
relevant to our work is that of contextual bandits where the algorithm
also has side information about the user's preferences.
%\cite{Langford07}.
%Excellent text book treatments on MAB algorithms are available
%in \cite{Bubeck12,Lattimore18,Slivkins19}. These references also
%describe many extensions and variations that have been developed on
%this basic model. In 

A key assumption made in the design and analyses of classical MAB
algorithms is that the reward distributions of the arms do not change
with time. This also implies that the preferences of the user
population are assumed to not be affected by the sequence of arms that
are played. From ones own experiences, the latter is clearly not
always true---user preferencess are affected by the
recommendations. In fact, the objective of advertisements is to
persuade users to adopt particular preferences. With this
motivation, in this paper, we make the reasonable assumption that user
preferences are not independent of the recommendations and are in fact
influenced by them. Specifically, we assume that the preferences of
the users at any time, and hence the reward distributions from the
arms, depends on the history of recommendations upto that time and is
a function of the arms that are played and rewards accrued.

MAB algorithms are usually designed to find the best arm, the one with
the highest expected reward, by a judicious combination of exploration
and exploitation. Our objective is a marked departure from this: the
goal is to actively use the preference reinforcements to influence the
population preferences. This objective is similar in spirit to the
emerging literature on opinion control,
e.g., \cite{Borkar15,Eshghi17,Goyal19}.  However, to the best of our
knowledge, such an objective has not been explored in an MAB setting.

In this paper, we will assume that the recommendation system uses a
contextual MAB algorithm to serve a multitype population with each
type having a unique preferred arm. We allow both positive and
negative influences to be simultaneously present---influence is
positive when the recommended arm is liked by the user and the
influence is negative when the recommendation is not liked by the
user. Furthermore, we consider two kinds of time-dependent behavior of
influence dynamics: (1)~decreasing intensity of influence where the
population becomes increasingly rigid in its preferences, and
(2)~constant intensity of influence where the degree of influence
remains constant throughout the period of interest. The first model is
motivated by the empirical observation that advertising gives
diminishing returns \cite{Palda65},\cite{Cowling10} and the second
model is inspired by the popular voter model (introduced
in \cite{Holley75}) which has been widely studied in the literature on
opinion dynamics. For both these models we will first consider a
system that is serving a population of two types with type 1 being the
preferred type, i.e., the objective of the algorithm is to maximize
the population of type 1 users at the ed of the time horizon $T.$

The following are the key contributions in the paper.

\begin{itemize}
\item
Optimal policy to maximize type~1 population when reward statistics
are known. Interestingly, the optimal policy does not necessarily
recommend the favoured arm. [Section~\ref{sec:known-B}]

\item
An explore-then-commit (ETC) policy to maximize type 1 population when
the reward statistics are unknown. This has to battle the twin
tradeoffs---usual exploration-exploitation tradeoff and the decreasing
ability to shape the preferences when the population has become more
rigid. This is also shown to have logarithmic regret in the known
time-horizon setting. [Section~\ref{sec:unknown-B}]

\item
A Thompson sampling based policy to maximize type 1 population when
the reward statistics are unknown. We show that this policy gives us
logarithmic regret even if the time-horizon is
unknown. [Section~\ref{sec:unknown-B}]

\item
For the case of the constant influence model, we show that the
analyses and the algorithms we obtained for the decreasing influence
model stay valid. [Section~\ref{sec:voter}]

\item
Finally, we present two different extensions, First, we describe an
$N$-arm MAB for an $N$-type population. For the objective of
maximising the type~1 population, we obtain the optimal policy when
the reward matrix is known. We then extend the Thompson sampling
algorithm to this case. This is considered in
Section~\ref{sec:general}.

A second extension is a model for two recommendation systems with
competing objectives for the two-arm, two-type case. Some preliminary
results are presented Section~\ref{sec:twoSys}.
\end{itemize}

In the next section we discuss the related literature and delineate
our work. The model and some preliminaries are set up in
Section~\ref{sec:model}. We conclude the paper with a discussion on
directions of future work. All the proofs are carried in the appendix.

\section{Related Work}
\label{sec:literature}

The literature on MAB algorithms is vast and varied and excellent
textbook treatments that describe the basic models and several key
variations are available in, e.g.,
\cite{Gittins11,Bubeck12,Lattimore18,Slivkins19}.
Our interest in this paper is on contextual bandits where the MAB has
side information about the user. An early analysis of MABs with side
information is in \cite{Wang05} although the term was coined
in \cite{Langford07}. In this paper we consider contextual bandits
where the type of the user is known to the algorithm before the
recommendation is made.

As we mentioned earlier, in the classical setting the reward
distributions on the arms remain the same and the rewards are
independently realized for each user. There is however, an emerging
literature on MAB algorithms that do not assume that reward
distributions are the same at all times. In \cite{Besbes19} the
expected reward from any arm can change at any point and any number of
times; changes are independent of the algorithm and the reward
sequence. In rotting and recharging bandits, the arms have memory of
when they were last recommended and the reward distribution depends on
the delay since the last use of the arm. In the rotting bandit
model \cite{Levine17}, the expected reward from playing an arm
decreases with the gap since the last play. The recharging bandit
model \cite{Immorlica18} is the opposite---expected reward increases
with the gap. The models of \cite{Meshram18,Meshram16} have similar
objectives but are developed in a Markovian restless bandits
setting. A generalisation of rotting and recharging bandits is
considered in \cite{Fiez19} where the mean rewards from the arms vary
according to a Markov chain. Positive reinforcement of the population
preferences based on the rewards seen is modelled in \cite{Shah18}.
Here the preference in a multi-type user population is positively
reinforced by positive rewards. However, this is not a contextual
bandit setting and negative reinforcements are not modelled here. 

None of the preceding are contextual bandits, i.e., they do not have
side information about the user.  Non stationary contextual bandits
are considered in \cite{Wu18} but the time dependent behaviour of the
arms is independent of the sample path of the user types and the
rewards.  In all of the preceding, the objective is to maximize the
cumulative reward (minimize regret) and not to influence the
population.

Our objective is to not maximize the rewards obtained but shape the
preferences. In this there are similarities with the objectives in
\cite{Borkar15,Eshghi17,Goyal19}. In \cite{Borkar15}, the objective is to
rapidly converge to a desired asymptotic consensus in a social
network. In \cite{Eshghi17,Goyal19}, open loop, optimal control
techniques are used to shape the opinions in a social network. We do
not deal with a social network, rather with the population that
interacts with a platform like a recommendation system. We believe
that this is the first work that models positive and negative
reinforcements due to the rewards and has the the explicit objective
of shaping the preferences in the population.

In the model that we describe in this paper, the preferences of the
user population will be tracked by a Poly\`{a} urn \cite{Polya30},
widely used to model random reinforcements; see \cite{Pemantle07} for
an excellent tutorial and survey. Much of this literature is on
characterization of the asymptotic composition of the urn. Our
interest in this paper is to influence the composition of the urn.

\section{Model and Preliminaries}
\label{sec:model}

A population of two types of {users} is served by a {recommendation
system} $S$ which recommends one of two arms to each arriving
user. The two types of users are distinguished by their preference for
one arm over another. Time is discrete and takes values $t \in [1:T].$
At time $t$, a user of type $X_t$ arrives, the type is observed by $S$
and the user is shown arm $A_t.$ Since the type is observed by $S$
before the arm recommendation is made, this is a contextual
bandit. For much of the paper we will assume $A_t \in
\{1,2\}$ and $X_t \in \{1,2\},$
i.e., we will consider a two-arm system serving a two-type population.

The probability of the incoming user at time $t$ being of a specific
type is proportional to the fraction of users of that type in the
population at time $t$. The fraction of users of a specific type in
the population evolves over time. Specifically, the fraction of users
in the population of a specific type at a given time is a function of
the recommendations made to incoming users up to that time and the
users' response to those recommendations.  For ease of exposition, we
track the fraction of type 1 and type 2 users in the population via
a \emph{virtual} urn containing colored balls (colors 1 and 2
corresponding to, respectively, population types 1 and 2); the
fraction of type 1 users in the population equals the fraction of type
1 balls in the urn.

%\color{red} [Old text] The fraction of type 1 and type 2 users in the
%population is tracked by an urn containing colored balls (colors 1
%and 2 corresponding to, respectively, population types 1 and 2); the
%fraction of type 1 users in the population equals the fraction of
%type 1 balls in the urn.\color{black}

\textbf{Reward Structure.}
Suppose $X_t = i$ and $A_t=j,$ i.e., user at time $t$ is of type $i$
and is recommended arm $j.$ The user gets a random Bernoulli reward
$W_t \in \{0,1\}$ with mean $b_{ij}.$ Let $B := [[b_{ij}]]$ be the
reward means matrix.  Without loss of generality, we assume that
$b_{ii}$ is the maximum in row $i$ of $B,$ i.e., type $i$ users prefer
arm $i$ over others.

\textbf{Population Dynamics.}
Let $Z_i(t)$ be the number of type $i$ balls in the urn at time $t.$
The user arriving at time $t$ is of type $i$ with probability $z_i(t)
= Z_i(t)/(\sum_j Z_j(t)).$ $W_t$ is the realization of the reward at
time $t$ and it causes the urn to be updated reflecting the change in
the population type effected by the arm shown and the reward
obtained. We consider two ways in which the update
occurs. $N_0=Z_1(0)+Z_2(0)$ is the total number of balls at $t=0.$
\begin{enumerate}
\item
\textbf{Decreasing Influence Dynamics (DID) model.}
In this model, the total number of balls increases by one in each
time-slot and the colour of the ball is determined as follows. If the
user arriving at time $t$ is shown arm $j,$ and the user likes this
arm (i.e., the random Bernoulli reward obtained is one) then the
number of balls of type $j,$ (i.e., the type that has preference for
the arm shown) increases by one. If the user does not like the arm
shown, then the number of balls of type ${-j},$ (i.e., the type that
has lower preference for the arm shown) increases by one. Formally,
\begin{align}
  Z_{A_t}(t+1)  &= Z_{A_t}(t) + W_t, \nonumber \\
  Z_{-A_t}(t+1) & = Z_{-A_t}(t) + (1-W_t).
  \label{eq:popdynam-dim}
\end{align}
Here $-A_t$ is the arm that was not recommended. The population
dynamics of \eqref{eq:popdynam-dim} implies that the maximum possible
change in the value of $z_{i}(t)$ decreases with $t.$ Thus the
population of users, and hence the preferences, becomes less plastic
with time.
%\DM{Need a motivating scenario? Generalization to adding more balls
%per timeslot?}
Real life examples of this setting are applications like Yelp which
recommend restaurants to customers. Customers review or rate
restaurants once they visit them. The ratings influence the
preferences of future customors. Furthermore, since the number of
reviews for each restaurant increase over time, the maximum possible
change to the average rating decreases with time.

\item
\textbf{Constant Influence Dynamics (CID) model .}  
In this model, the total number of balls in the urn remains constant
over time. The influence of the rewards over the population is modeled
as follows.  If a user of type $X_t$ arriving at $t$ yields reward 1
when shown arm $A_t={-X_t}$ or reward 0 when shown arm $A_t={X_t}$
then one ball of type $-X_t$ changes its colour. The balls do not
change type in the other two cases. Formally, defining $\theta_t$ to
be the event $\{A_t={-X_t}\},$ the urn evolution will be
\begin{align}
  Z_{A_t}(t+1) &= Z_{A_t}(t) + \left( \mathbbm{1}_{\theta_t} \oplus
  W_t\right), \nonumber \\
  Z_{-A_t}(t+1) & = Z_{-A_t}(t) - \left( \mathbbm{1}_{\theta_t} \oplus
  W_t\right).
\label{eq:popdynam-cim}
\end{align}
Here the maximum possible change in the value of $z_{i}(t)$ remains
constant over time. This model is inspired by the voter model
of \cite{Holley75} which is usually defined on a graph. Here the
voters interact with the recommendation system, one at a time, rather
than with each other. Real life examples of this setting include
tracking polls that use a focus group of the same set of individuals
that are polled periodically to track the evolution of their
collective opinion over time.
\end{enumerate}

%\DM{Need a motivating scenario? add probability of switching?}

Observe that in both the models, the rewards are stationary, i.e.,
$b_{ij}$ does not chage with time. However, the preferences for the
arms in the population, $z(t),$ is changing with time. Thus this is
a \textit{non stationary contextual multi-armed bandit.}

\textbf{Remark.} It is easy to see that both the models are a special
case of the following general Markov decision process model. Let
$(Z(t), X_t)$ be the state of the system with $X_t$ being a function
of $Z(t).$ The evolution of $Z(t+1),$ and hence $X_{t+1},$ depends on
the action $A_t$ \emph{and} on the reward $W_t.$

\DM{%Check this. To conform with the definition of regret, I have made
    %it a trajectory objective.
    We only prove $\lim_{T \to \infty} x(T)$ is
    maximised. Right?}
    \VN{Yes}

\textbf{Algorithmic Goal.} The goal of the recommendation system $S$ 
is to follow a trajectory $z(t)$ for $0 < t \leq T$ that achieves the
maximum possible increase at at every time $t.$ We will show in
Theorem~\ref{thm:propmax} that this also corresponds to maximizing
$z(T)$ as $T\to \infty.$

We now formally define a policy.% and then an optimal policy. 

\DM{Should the policy also include the history and not just $z_i(t)$?}
\VN{Address this}

\begin{defn}[Policy]
  \label{def:policy} A {policy} $\pi$ is a time indexed sequence of
  the tuples $(p_t,q_t)$ where, for all $t \in [1:T],$
  \begin{align}
    \nonumber
    p_t & = P(A_t = a_1| X_t = 1, z_1(t))\\
    q_t & = P(A_t = a_2| X_t = 2, z_1(t)).
    \label{eq:policy}
  \end{align}
\end{defn}
From the definition, $p_t$ is the probability of showing arm 1 to a
type 1 user and $q_t$ is the probability of showing arm 2 to a type 2
user. The sequence of tuples $\{p_t, q_t\}_{t>0}$ uniquely identifies
a policy for the two-armed influencing bandit problem. We now
formalize our goal by defining the notion of an optimal policy.

%\DM{VN: You probably meant to say this}
%From the preceding discussion, we see that $\{X_{\tau}, A_{\tau},
%W_{\tau}\}_{\tau < t} \ \rightarrow Z(t) \rightarrow z(t)$ is Markov
%chain, i.e., given $Z(t),$ $z(t)$ is conditionally independent of
%$\{X_{\tau}, A_{\tau}, W_{\tau}\}_{\tau < t}.$ Thus 
%
%
%
%The pmf of 
%$Z(t) \}$ is the same as that of $\{X_t \ | \ Z(t)\},$ i.e., $\{X_t,
%Z(t) \}$ is a Markov chain. Hence, we define a locally optimal policy
%conditioned only on $z(t)$ instead of on the history, $\{X_{\tau},
%A_{\tau}, W_{\tau}\}_{\tau < t}.$

\begin{defn}[Optimal policy]
  \label{def:optimal} An optimal policy for time $t$ is the value of
  $(p_t,q_t)$ that maximizes the expected increase in type 1
  population in time slot $t,$ given the population profile at time
  $t,$ i.e.,
  \begin{equation}
    (p^{*}_t,q_t^{*}) = \arg \max_{(p_t,q_t)} \EXP{ \Delta
      Z_1(t) \ | \ z_1(t)} \label{eq:opt-policy}
  \end{equation}
  where $\Delta Z_1(t) = Z_1(t+1) - Z_1(t)$.
\end{defn}
%
%\DM{Revisit the following remark}

\textbf{Remark.} Observe that this definition is of a locally optimal
policy. However, in Theorem~\ref{thm:propmax} of the next section we
show that the optimal policy is independent of $z_1(t)$ and thus the
locally optimal policy is also optimal in a broader sense that is made
more explicit.

Next we define the metrics of regret and cumulative regret for preference
shaping which will be used to compare the performance of different
policies. Let $\Delta Z_1^{\pi}(t)$ (resp. $\Delta Z_1^{*}(t)$) be the
change in the population of type 1 balls in the urn at time $t$ given
that we follow the policy $\pi$ (resp. the optimal policy) at time
$t.$ 

\begin{defn}[One-step regret]
  \label{def:regret} Regret in a time slot $t,$ for policy $\pi,$
  denoted by $R_t^{\pi},$ is defined as
  \begin{equation}
    R_t^{\pi} = \EXP{ \Delta Z_1^{*}(t) - \Delta Z_1^{\pi}(t)\ |
      \ Z_1^{*}(t) = Z_1^{\pi}(t) }.
     \label{eq:regret}
  \end{equation}
  \end{defn}

The definition of cumulative regret follows naturally.

\begin{defn}[Cumulative regret] 
  \label{totreg}
  {Cumulative regret} ($R_{[1:T]}^{\pi}$) of a policy $\pi$ is defined
  as
  \begin{displaymath}
    R_{[1:T]}^{\pi} = \sum_{t = 1}^{T} R_t^{\pi}.
  \end{displaymath}
\end{defn}

%\DM{The following needs more attention; it is a bit patchy at this time}
%\Rev{VN : Shift commonly defined regret discussion to appendix? }
Observe that in the definition of one-step regret, the expectation is
conditioned on the population profile $Z_1(t)$ for both the optimal
policy and candidate policy $\pi$ being the same at time $t^{-},$
before the policy is applied. The cumulative regret is just the sum of
these one-step deviations.
%Clearly, this is different from the definition of regret in most
%other MAB analyses where the regret is unconditional. 

We argue that a meaningful comparison between two policies in a time
slot can be done when both are operating on the same composition of
population. Hence, when defining one-step (or instantaneous) regret,
we specify the population composition on which the optimal policy is
being played. Such a metric is a useful, and informative, to compare
policies, because if policy $A$ is better than another policy $B$ in
maximizing the expected increase in type 1 balls in the urn at all
time instants conditioned on the same initial state, then it will also
incur a lesser regret for all time instants.

%\DM{The following does not appear to be complete; may be we should
%discuss to see what is appropriate here. Revise based on Lemma 7}

Adapting the commonly used definition of regret for our setting would
define the regret of a candidate policy as the difference between the
maximum possible value of $z_T$ and the value of $z_T$ under a
candidate policy. We make a precise comparison between this commonly
used definition and our definition in \eqref{eq:regret} in
Appendix \ref{sec:regComp}.

\DM{I am still not fully comfortable with this section. Please reread and
we can discuss, specifically, the notion of regret.}
\VN{The only reason behind defining regret this way was ease of analysis and the fact that stepwise optimal policy was also asymptotically optimal. }

%Seems to me that two finite $T$ comparisons are possible to
%characterize regret: (1)~Compare with maximum possible $z_T.$
%(2)~Compare with maximum possible $\left( \sum_{t=1}^T z_t \right).$
%Let us discuss 
    
%The regret defined in \eqref{eq:regret} is a weaker metric than the
%both of these

%With the usual definition would be to compare the policies at the end
%the time horizon $T$ and measure the population fraction (since it is
%a one-way ``if'' as opposed to ``if and only if'').

%We make a more precise comparison between the two definitions of
%regret in Appendix \ref{sec:regComp}.

\section{Preference Shaping with Known Rewards Matrix}
\label{sec:known-B}
In this and in the next section we consider the DID model in which the
influence of the actions of $S$ and the observed rewards decreases
with time. We begin by assuming that $S$ knows the reward means matrix
$B$ and wants to maximize the expected proportion of type 1 users in
the population at time $T.$ This case is interesting in its own right
because it is directly useful when $B$ is exogenously available, e.g.,
from previously collected data. It also provides insight into the
behavior of the optimal influence process. Thus it is a preliminary
result that informs the design of algorithms for the case when $B$ is
unknown.

%The optimal policy for this case turns out to be stationary and has a
%simple form.

The following theorem describes the optimal policy. 
\begin{thm}
  \label{lem:opt1}
  The optimal policy at time $t$ is 
  \begin{equation}
    (p_t^{*},q_t^{*}) = (\mathbbm{1}_{\{ b_{11} + b_{12} - 1 >
      0\}},\mathbbm{1}_{\{ b_{21} + b_{22} - 1 < 0\}})
    \label{eq:policy-known-B}
  \end{equation}
  where $\mathbbm{1}_{\{\cdot \}}$ is the indicator function.
\end{thm}

The optimal policy would have $S$ recommending arm $1$ to a type 1
arrival for all $t$ if $b_{11} > 1 - b_{12}$ and recommending arm $2$
otherwise. Recall that arm $1$ is the preferred arm for type 1
arrivals. We can interpret this to mean that $S$ should always
recommend arm $1$ to type 1 users if they like arm $1$ more than they
dislike arm $2,$ and should recommend arm $2$ otherwise. This is a
result of the negative reinforcement that can happen if arm $1$ is not
liked sufficiently strongly or if arm $2$ is liked strongly.
%Similarly for type 2 arrivals.
Thus, for optimal preference shaping, the system \textit{may recommend
  arms which have a lower preferrence for the user.}
%For example, if the matrix $B = (b_{11}=0.9,b_{12}=0.3,b_{21} =
%0.4,b_{22} = 0.7)$, then the optimal policy is $(p^*=1,q^*=0)$.

Clearly, the policy of \eqref{eq:policy-known-B} in
Theorem~\ref{lem:opt1} gives us zero cumulative regret. However, it is
not immediately clear if this policy maximizes the proportion of type~1
users in the population at time $T.$ To show that this is indeed the
case, we first analyze the time evolution of the expected proportion
for policy $\pi$ and state the following lemma.
\DM{We only show asymptotic properties; why would that also maximize
at $T?$ }
\VN{Yes it is not necessary, should change this to large $T$ or $T \to \infty$}

\begin{lem}
  \label{lem:prop1}

  Let $d_1 := p(1-b_{11}) + (1-p)b_{12}$ and $d_2 := q(1-b_{22}) +
  (1-q)b_{21}.$ Further, let $z_1(0)$ be the proportion of type 1
  users at $t=0.$ For a policy $\pi$ with $(p_t,q_t) = (p,q),$ the
  expected proportion of type 1 users at time $t$ is
  \begin{equation}
    z_1(t) = \frac{d_2}{d_1 + d_2} + \left( z_1(0) -
    \frac{d_2}{d_1 + d_2}\right) \left(1
      + \frac{t}{N_0} \right)^{-(d_1 + d_2)}
      \label{eq:DI-z1-of-t}
    \end{equation}

\end{lem}

Lemma \ref{lem:prop1} tells us that the expected proportion of type 1
users monotonically approaches $d_2/(d_1 + d_2)$ as $t \to \infty$ and
is independent of $z_1(0).$ Thus, this is the maximum proportion that
can be achieved using a policy of the form $(p_t=p,q_t=q)$. It turns
out that the policy from \eqref{eq:policy-known-B} of
Lemma~\ref{lem:opt1} does indeed maximize the proportion of type 1
users in the population.
\begin{thm}
  \label{thm:propmax}

  The optimal policy of \eqref{eq:policy-known-B} of
  Lemma~\ref{lem:opt1} maximizes the expected asymptotic proportion of
  type 1 users and the maximum proportion is $\left( \frac{d_2}{d_1 +
  d_2}\right).$
 
\end{thm}

Thus, if the matrix of mean rewards $B$ is known, we can obtain a zero
regret policy that also provably maximizes the expected proportion for
the goal of preference shaping.

The preceding results are aimed at optimizing metrics that are in
expectation. We can say more; specifically, we can provide long and
short term sample path guarantees via stochastic approximation
theory. Furthermore, the derivation of the mean trajectories as in
Lemma~\ref{lem:prop1} through the formation of an o.d.e can also be
justified using this theory. We do this below.

\textbf{Sample path guarantees.} Recall that the classic stochastic
approximation equation for which we have long term convergence
guarantees is the following equation, introduced by Robbins and Monro
in \cite{Robbins51}
$$x_{t+1} = x_t + a(t)(h(x_t) + M_{t+1}).$$
Here $h$ is Lipschitz, $M_t$ is a zero-mean martingale difference
sequence and $\{a(t)\}$ are coefficients which satisfy the two
conditions (1)~$\sum_t a(t) = \infty$ and (2)~$\sum_t a^2(t) <
\infty$. For such an iteration, it is guaranteed that all sample paths
of $x_t$ will converge to the equilibrium point of the
o.d.e. $\Dot{x}(t) = h(x(t))$ (i.e. $h(x) = 0$) almost surely as $t
\to \infty.$ Further, the maximum deviation from the o.d.e. trajectory
in a fixed length time window $[t,t+T]$ also goes to 0 almost surely
as t goes to $\infty$. A weaker error bound for finite time and
constant stepsize $a(t)=a$ is given in \cite{Kumar182}. \DM{Should we
state the actual result adapted to this problem?} \VN{would be useful for only B case}

We now analyze the difference between the trajectory of the
o.d.e. obtained in the proof of Lemma~\ref{lem:prop1} and a sample
path of the stochastic sequence evolving according to
\eqref{eq:popdynam-dim} using stochastic approximation theory. Towards
this goal, we first rewrite \eqref{eq:popdynam-dim} into the classical
Robbins-Monro form. For a fixed policy, the stochastic sequence of the
number of type 1 balls, $Z_1(t),$ evolves as
\begin{align*}
  Z_1(t+1) &= Z_1(t) + \Delta Z_1(t)\\
  &= Z_1(t) + \left( E[\Delta Z_1(t)|z_1(t)] \right. \\
  & \hspace{25pt} \left. + (\Delta Z_1(t) -E[\Delta Z_1(t)|z_1(t)])
  \right) ,
\end{align*} 
where $ E[\Delta Z_1(t)|z_1(t)]= z_1(t)(1-d_1) + (1 - z_1(t))d_2$
(which can be shown directly from the definition of the DID
model). Since $E[\Delta Z_1(t)|z_1(t)]$ is Lipschitz in $Z_1(t)$ and
$M_t=\Delta Z_1(t)-E[\Delta Z_1(t)|z_1(t)]$ is a zero mean martingale
difference sequence, this equation is in a constant stepsize form of
the Robbins-Monro iteration and has been used to obtain the o.d.e. in
the proof of Lemma~\ref{lem:prop1}. To get long term convergence
guarantees, we divide the above equation by $(N_0 + t + 1)$ on both
sides to obtain the following. 
\begin{align*}
z_1(t+1) &= \frac{1}{N_0 + t + 1} \left( Z_1(t) + E[\Delta
    Z_1(t)|z_1(t)] +M_t \right)\\  
  & \hspace{-30pt}= \frac{N_0+t }{N_0 + t + 1}z_1(t) + \frac{1}{N_0 +
    t + 1}( E[\Delta Z_1(t)|z_1(t)] + M_t)\\
  &\hspace{-30pt} = z_1(t) + \frac{1}{N_0 + t + 1}( E[\Delta
    Z_1(t)|z_1(t)] - z_1(t) + M_t)\\
  &\hspace{-30pt} =z_1(t) + \frac{1}{N_0 + t + 1}( d_2 -
  (d_1+d_2)z_1(t) + M_t)
\end{align*}
In the preceding we have used $Z_1(t+1)/(N_0+t+1) = z_1(t+1).$ This
last equation is now in the standard Robbins-Monro form as in
\cite{Robbins51}, since $\sum_t 1/(N_0+t+1) = \infty$ and $\sum_t
1/(N_0+t+1)^2 < \infty.$ Therefore, all sample paths followed by
$z_1(t)$ converge asymptotically almost surely to $z_1 =
d_2/(d_1+d_2).$

\begin{figure}
  \begin{center}
    \includegraphics[width=3.4in]{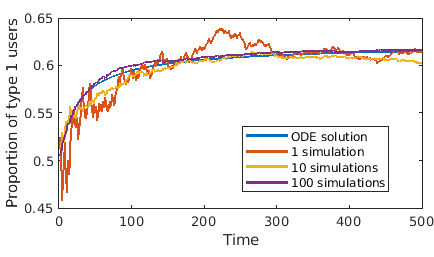}
  \end{center}
  \caption{Trajectory of proportion of type 1 population. Comparison
    of o.d.e. solution and averages from 1, 10 amd 100 sample
    paths. We used $B$ with the values $b_{00}=0.9,$ $b_{01}=0.4,$
    $b_{10}=0.2,$ and $b_{11}=0.6.$
    \label{fig:odesimcomp}
  }
\end{figure}

Figure~\ref{fig:odesimcomp} shows an example comparison between the
trajectories from 1, 10, and 100 samples paths. The figure also plots
the trajectory of the o.d.e. obtained in Lemma
\ref{lem:prop1}. Observe that although a single sample path can have
substantial deviations, averages of even a small number begins to
track the o.d.e. rather closely. This observation was seen in all of
our many simulations.

%\color{blue}
%show comparison plots for simulated and analytically plotted? Do for
%increasing number of simulations to see how much the sample paths
%deviated from ode's expected solution
%\color{black}

\section{Preference Shaping with Unknown Rewards Matrix}
\label{sec:unknown-B}
When the rewards mean matrix $B$ is unknown, we face a two-fold
trade-off in preference shaping. The first trade-off is the classical
exploration-exploitation trade-off, where exploration involves
correctly estimating the matrix $B$ and exploitation involves using
the optimal policy derived in the previous section to maximise the
fraction of type 1 population. The second trade-off is the decreasing
plasticity (DID model) of the preferences modelled by the increasing
number of balls in the urn. This puts additional pressure on
exploitation of the estimated $B$ as soon as possible.

We consider two algorithms to address the two-fold trade-off. We first
describe and analyze a naive explore-then-commit (ETC) algorithm. Next
we describe a Thompson sampling (TS) based algorithm and also analyze
it. We will see that if the time horizon $T$ is known, ETC can give
logarithmic regret. The TS algorithm also gives logarithmic regret and
has the advantage of not needing to know $T$ in the parametrization of
the algorithm.

Our analyses of the algorithms will obtain upper bounds on the
cumulative regret accrued by each algorithm. First we obtain the
regret for a general policy $\pi = \{p_t,q_t\}_{t>0}$ in the following
lemma.
\begin{lem}
  \label{lem:regexp}
  Define $\Delta_1 := |b_{11} + b_{12} - 1|$ and $\Delta_2 := |b_{22}
  + b_{21} - 1|$. The regret $R_t^{\pi}$ of a policy $\pi$ that has
  parameter values $(p_t,q_t)$ is given by
  \begin{align*}
    R_t^{\pi} &= z_1(t)|\mathbbm{1}_{\{ b_{11} + b_{12} - 1 > 0\}} -
    p_t|\Delta_1\\
    & \hspace{0.5in} + (1-z_1(t))|\mathbbm{1}_{\{ b_{21} +
      b_{22} - 1 < 0\}} - q_t|\Delta_2 .
  \end{align*}
\end{lem}

\subsection{The Explore-Then-Commit (ETC) Algorithm}
This algorithm has two phases. The exploration phase lasts for $m$
time units when each arm is recommended uniformly and the rewards mean
matrix is estimated. This estimate is used to determine the optimal
policy and commit to it during the remaining $(T-m)$ time units. Based
on the preceding section and assuming that the estimates then use the
optimal policy of the previous section (evaluated for the estimate
$\hat{B}= [\hat{b}_{ij}]_{2\times2}$) to maximize the proportion of
type 1 users. Algorithm~\ref{algo:ETC} describes the scheme in detail.

\begin{algorithm}
  \SetAlgoLined
  %\KwResult{Write here the result }
  Initialize $b_{ij} = 0$, $n_{ij} = 1$ for all $i,j \in \{1,2\}$\;
  \For{t = [1:m]}{
    $i \gets $Type of user arrived\;
    Show arm $j$ chosen uniformly from $\{1,2\}$\;
    Collect reward $W_t$\;
    $b_{ij} += W_t$\;
    $n_{ij} += 1$\;
    %   \eIf{condition}{
    %   instructions1\;
    %   instructions2\;
    %   }{
    %   instructions3\;
    %   }
  }
  $\hat{b}_{ij} \gets b_{ij}/n_{ij}$ for all $i,j \in \{1,2\}$\;
  \For{t = [m+1:T]}{
    $i \gets $Type of user arrived\;
    \eIf{i == 1}{
      Show arm 1 w.p $\mathbbm{1}_{\{\hat{b}_{11} + \hat{b}_{12} - 1 > 0\}}$\, else show arm 2;
    }{
      Show arm 2 w.p $\mathbbm{1}_{\{\hat{b}_{22} + \hat{b}_{21} - 1 < 0\}}$\ else show arm 1;  
    }
  }
  
  \caption{The Explore-Then-Commit algorithm}
  \label{algo:ETC}
\end{algorithm}

We now use Lemma~\ref{lem:regexp} to find the regret for this policy.

\begin{lem}
\label{lem:regexpetc}
  The cumulative regret, for the Explore-then-Commit policy with $m$
  time units for exploring is
  \begin{align*}
    R_{[1:T]}^{\pi} &= R_{explore} + R_{commit}, 
  \end{align*}
  where
  \begin{align*} 
    R_{explore} &= 0.5\left(\sum_{t = 1}^{m}z_1(t)\right)\Delta_1 +
      0.5\left(m-\sum_{t = 1}^{m}z_1(t)\right)\Delta_2\\
    R_{commit} &= p_{err}\left(\sum_{t = m+1}^{T}z_1(t)\right)\Delta_1 \\
    & \hspace{0.3in} + q_{err}\left( T-m - \sum_{t =
    m+1}^{T}z_1(t) \right)\Delta_2\\
    p_{err} &= P((\hat{b}_{11} + \hat{b}_{12} - 1)(b_{11} + b_{12} -
    1) < 0)\\ 
    q_{err} &= P((\hat{b}_{22} + \hat{b}_{21} - 1)(b_{22} + b_{21} - 1) <
    0)
\end{align*}
\end{lem}

It now remains to derive bounds on the cumulative regret so that this
policy can be compared to others. The general result has eluded
us. However, we state the following result for the the special case of
$b_{11} = b_{22}$ and $b_{12} = b_{21}.$

\begin{thm}
  \label{ETC}
  If $B$ is such that $b_{11} = b_{22}$ and $b_{12} = b_{21},$ the
  cumulative regret for the Explore-Then-Commit (ETC) policy is
  bounded above by           
  \begin{equation}\label{eq:etcBound1}
    R_{[1:T]}^{ETC} \leq m\Delta_1/2 + (T-m)\Delta_1e^{-m\Delta_1^2/8}
  \end{equation}

  Further, using $m = 8 \log(T)/\Delta_1^2$ (to bound the regret in
  terms of $T$ and eliminate $m$), we get a logarithmic regret, i.e.,

  \begin{equation}\label{eq:etcBound2}
    R_{ETC} \leq \frac{4}{\Delta_1} \log(T) + \mathcal{O}(1/T).
  \end{equation}
\end{thm}

Thus for a finite time horizon for some $B,$ we can indeed get
logarithmic regret. A situation obeying the conditions specified in the previous theorem has been shown in Fig.~\ref{popvtimesym}.

The performance of the general ETC is open; the optimal $m$ is not
known. Even for the special case, $m$ requires $T$ to be
known. Furthermore, the ETC algorithm is inherently inefficient
because it has to spend a significant amount of time in the initial
slots, when the preferences are more plastic (equivalently, the
influence of the rewards is higher), doing exploration to estimate
$B.$ Both of these drawbacks suggest that a better policy could be to
estimate $B$ as well as track a confidence level of that estimate,
which would tell us whether to explore or not. We outline such a policy next.

\subsection{Thompson Sampling}
 The Thompson sampling algorithm that we present in Algorithm~\ref{algo:TS} seeks to overcome the drawbacks of the ETC policy.

Algorithm~\ref{algo:TS} maintains a prior on the $b_{ij}$ and in each
time slot, values are updates it based on the reward obtained. The estimate for
$b_{ij}$ in every time slot is sampled by a Beta distribution that decreases its variance with every new sample obtained. This automatically takes care of the exploration-exploitation trade-off.

\begin{algorithm}
  \SetAlgoLined
  %\KwResult{Write here the result }
  Initialize $\alpha_{ij} = 1$, $\beta_{ij} = 1$ for all $i,j \in \{1,2\}$\;
  
  \For{t = [1:T]}{
    $i \gets $Type of user arrived\;
    Sample $\Tilde{b}_{ij} \sim Beta(\alpha_{ij},\beta_{ij})$ for all $i,j \in \{1,2\}$\;
    \eIf{i == 1}{
      Show arm 1 w.p $\mathbbm{1}_{\{\Tilde{b}_{11} + \Tilde{b}_{12} - 1 > 0\}}$, else show arm 2;
    }{
      Show arm 2 w.p $\mathbbm{1}_{\{\Tilde{b}_{22} + \Tilde{b}_{21} - 1 < 0\}}$, else show arm 1;  
    }
    $j \gets$ Arm showed\;
    $R_t \gets$ Reward obtained\;
    $\alpha_{ij} \gets \alpha_{ij} + R_t$\;
    $\beta_{ij} \gets \beta_{ij} + (1-R_t)$\;
  }
  
  \caption{The Thompson sampling algorithm.}
  \label{algo:TS}
\end{algorithm}

The following theorem shows that the Thompson sampling policy can
provide logarithmic regret in general.

\begin{thm}
  The cumulative regret for the Thompson sampling policy is bounded
  above by
  \begin{equation}
    \label{eq:tsBound1}
    R_{[1:T]}^{Thomp} \leq \frac{(z^{*})^2}{4}\left(
     \frac{1}{f_1(1-f_1)\Delta_1} +
    \frac{1}{f_2(1-f_2)\Delta_2}\right) \log(T). 
  \end{equation}
  Here $z^{*}$ is the asymptotic proportion reached by the optimal
  policy for the matrix $B,$ and $f_1,f_2 < 1$ are constants that depend 
  on the parameters of the Thompson sampling procedure.
  \label{thm:Thompson1}
\end{thm}

\subsection{Simulations}

We have performed extensive simulations to study the performance of
the two policies. In all of these, we see that the Thompson sampling
policy far outperforms ETC in both maximizing the required population
proportion and minimizing the cumulative regret. Some representative
results are presented below. The details of the simulations are specified in Appendix \ref{sec:simDetails}.

Figure~\ref{popvtime1} shows the evolution of the proportion of type 1
population as a function of time for the ETC, TS, and the optimal
policies for various values of $B.$ To obtain the plot for the ETC
algorithm we tried all values of $m,$ $1 \leq m \leq T$ and the plot
for the best choice of $m,$ i.e., the $m$ that maximizes type 1
population at the $T,$ is shown.  The plot for the optimal policy
assumes $B$ is known and uses the policy
from \eqref{eq:policy-known-B} of Lemma~\ref{lem:opt1}.

\begin{figure}[!htbp]
  \begin{center}
  \includegraphics[width=3.40 in]{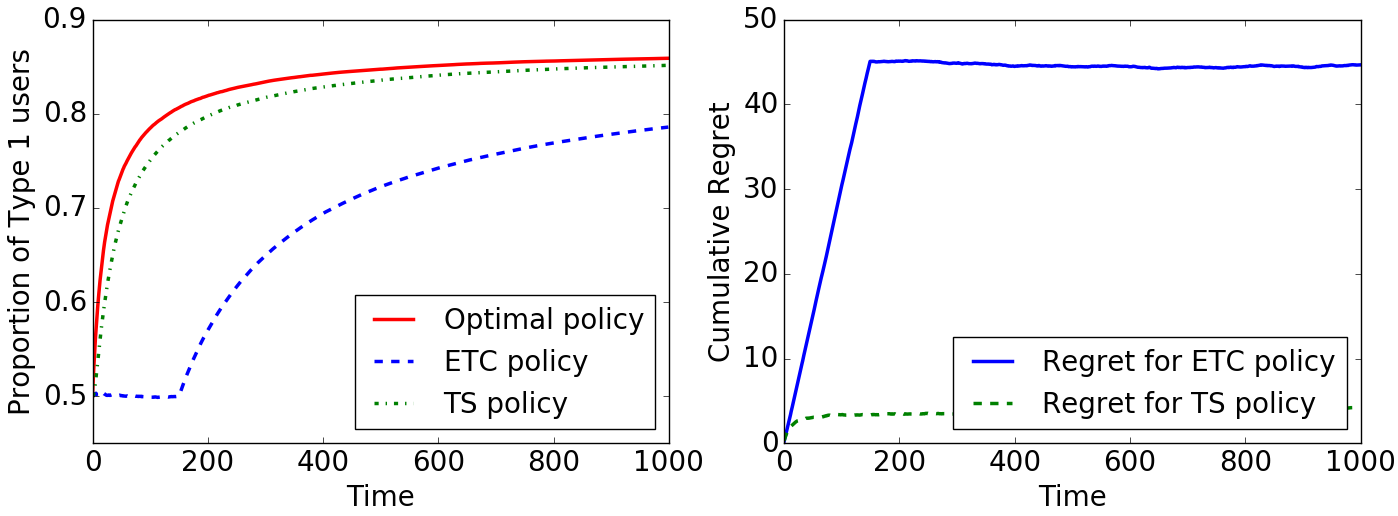}
  \end{center}

  \caption{ Expected population proportion vs time (left) and
  cumulative regret vs time (right) for the ETC, TS, and the optimal
  for $B_{sym} = (b_{11}=0.9, b_{12}=0.7, b_{21}=0.7, b_{22}=0.9)$}

  \label{popvtimesym}

\end{figure}

In all the cases, the TS policy does significantly better. In
Fig.~\ref{popvtimesym}, we consider the symmetric $B_{sym}=
(b_{11}=0.9, b_{12}=0.7, b_{21}=0.7, b_{22}=0.9)$ for which
Theorem~\ref{ETC} is applicable and we have a prescribed optimum
exploration duration $m$ for logarithmic regret. We see that even here
the TS scheme does significantly better than ETC.

\begin{figure}
      
  \begin{center}
    \begin{subfigure}{\columnwidth}
    \begin{center}
    \includegraphics[width=3.40 in]{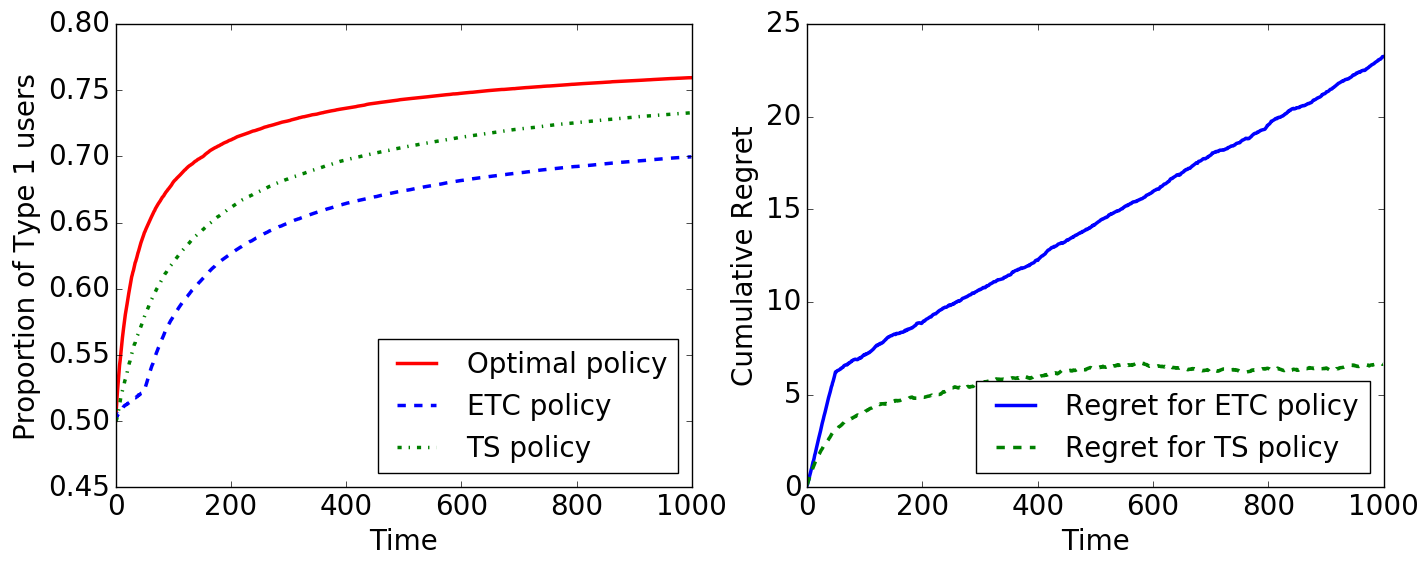}

    \caption{$B_1 = (b_{11}=0.9, b_{12}=0.4, b_{21}=0.2, b_{22}=0.6).$
    Optimal policy is $(p=1,q=1)$ with asymptotic $z_1 = 0.80$.  }
    \end{center}
    \end{subfigure}

    \begin{subfigure}{\columnwidth}
    \begin{center}
    
     \includegraphics[width=3.40in]{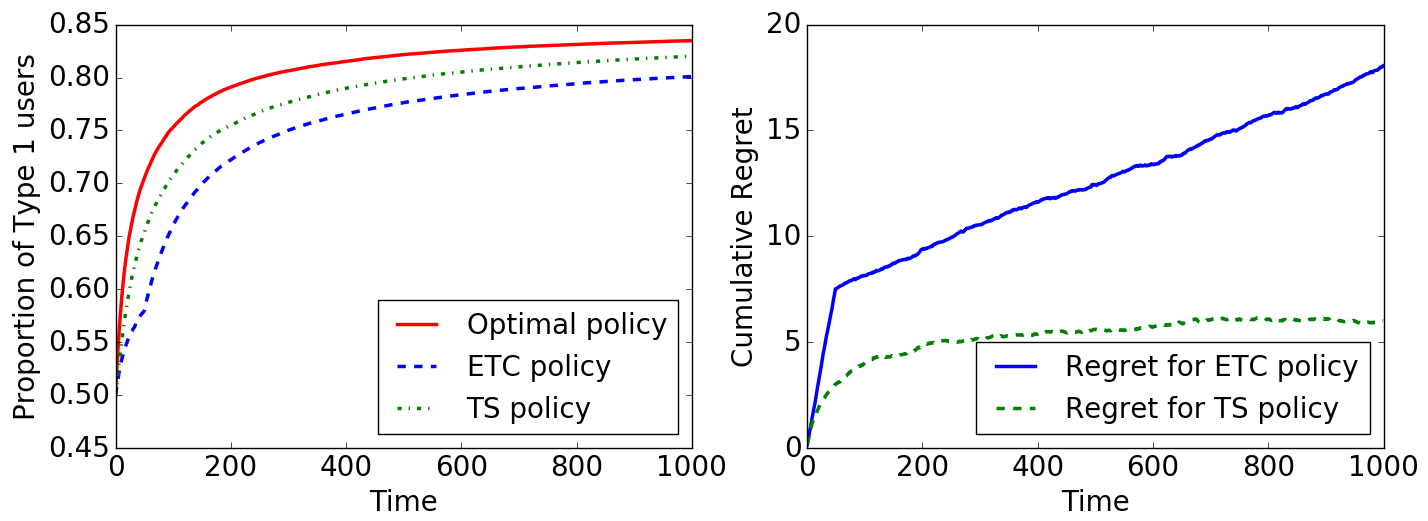}

    \caption{$B_2 = (b_{11}=0.9, b_{12}=0.4, b_{21}=0.6, b_{22}=0.7).$
    Optimal policy is $(p=1,q=0)$ with asymptotic $z_1 = 0.86$.}
    \end{center}
    \end{subfigure}
    
   \begin{subfigure}{\columnwidth}
   \begin{center}
 
   \includegraphics[width=3.40 in]{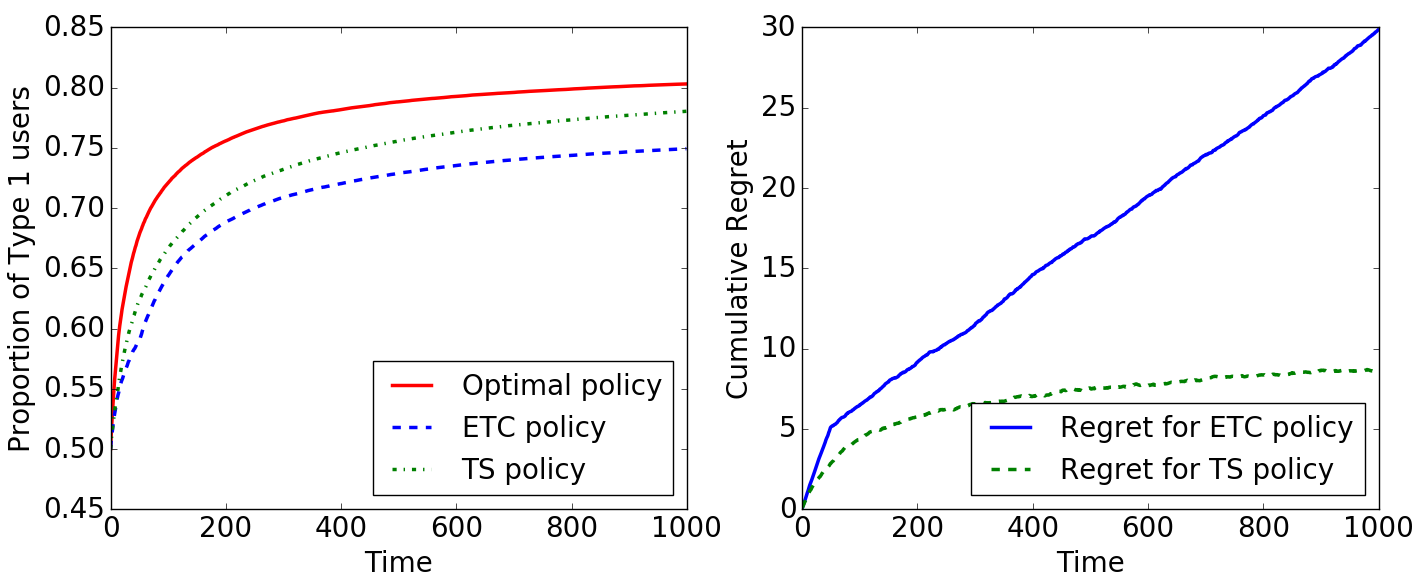}

   \caption{$B_3 = (b_{11}=0.7, b_{12}=0.1, b_{21}=0.3, b_{22}=0.5).$ Optimal
     policy is $(p=0,q=1)$ with asymptotic $z_1 = 0.83$.}
     \end{center}
     \end{subfigure}

    \begin{subfigure}{\columnwidth}
    \begin{center}

     \includegraphics[width=3.40 in]{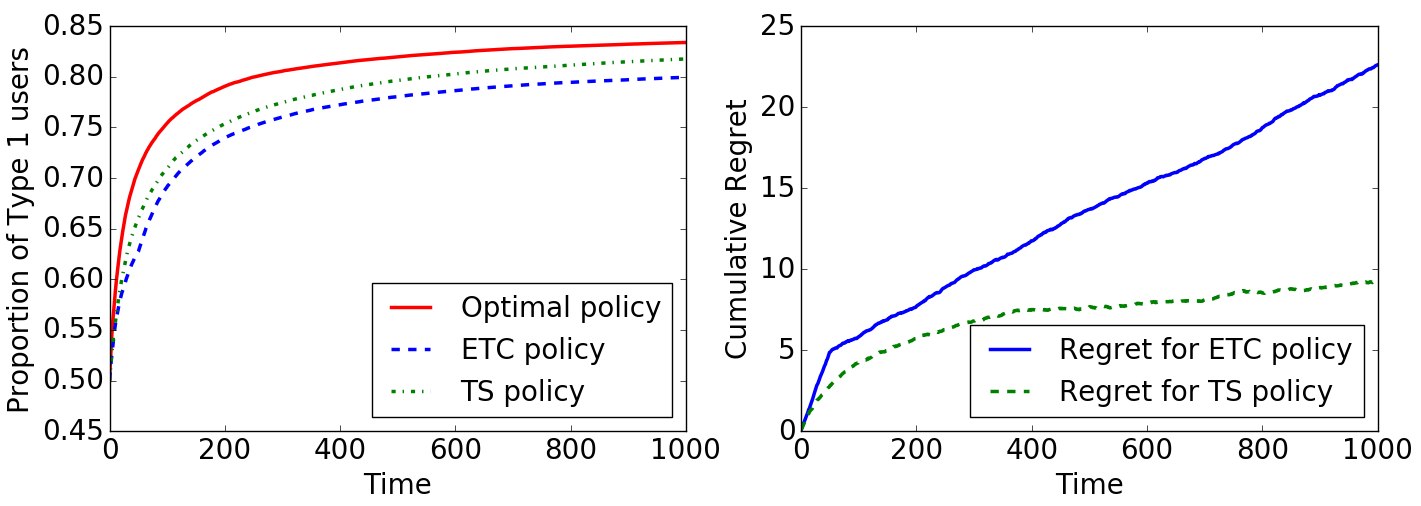}

     \caption{$B_4 = (b_{11}=0.7, b_{12}=0.1, b_{21}=0.6, b_{22}=0.6).$ Optimal
     policy is $(p=0,q=0)$ with asymptotic $z_1 = 0.86$.}
     \end{center}
     \end{subfigure}
  \end{center}

  \caption{Expected population proportion vs time (left) and
  cumulative regret vs time (right) for the ETC, TS, and the optimal
  policy that knows $B.$ The $B$ used for each of the plots and the
  optimal policy are also shown.} \label{popvtime1}
\end{figure}

\DM{Should we also plot the bounds to compare simulations}

%\DM{More details on how the simulations were done?}\VN{}

\DM{Should we have a short discussion to end the section?}

\section{Preference Shaping with Constant Influence}
\label{sec:voter}
%We also considered a model where the number of balls in the urn is
%fixed and the dynamics inspired by the voter model studied in opinion
%shaping problems.
In this section, we focus on the CID model in which the influence of
the rewards on the population preferences is constant over
time. Recall that the basic setup and reward structure remain the same
as that of the DID model. However, the impact of the reward, $W_t,$
accrued in each time slot changes. If a user of type $i$ arrives and
gets a unit reward when $A_t = -i,$ or if it gets reward 0 if $A_t =
i$ then one ball of type $i$ changes its color to the other color,
$-i.$ The composition of the urn remains unchanged in the other two
cases. Clearly, the ETC and Thompson sampling algorithms of,
respectively, Algorithm~\ref{algo:ETC} and Algorithm~\ref{algo:TS},
can be used for the CID model without any change. In the following we
will analyze their performance.

We first present the counterpart to Theorem~\ref{lem:opt1}. Assuming
that $B$ is known, the trajectory of the fraction of type 1 population
for policy $(p_t,q_t)=(p,q)$ is given by the following lemma.

\begin{thm}
  \label{lem:fixpopvt}
  For the CID model,for a policy $\pi$ such that $(p_t,q_t) = (p,q)$,
  the time evolution of the expected fraction of type 1 users is given
  by
  \begin{equation}
    z_1(t) = \frac{d_2}{d_1 + d_2} + \left(z_1^0 - \frac{d_2}{d_1 +
    d_2}\right)e^{-t\frac{d_1+d_2}{N_0}}.
    \label{eq:CI-z1-of-t}
  \end{equation}
  Here $d_1 = p(1-b_{11}) + (1-p)b_{12}$, $d_2 = q(1-b_{22}) +
  (1-q)b_{21}$ and $z_1(0)$ is the initial proportion of type 1 users.
\end{thm}

We see that, with a fixed $(p,q),$ the asymptotic fraction of type 1
in the population for the CID model has the same value as that of the
DID model. The difference though is in the rate at which the asymptotic
value is approached. Since the expressions for the asymptotic proportions for CID and DID are the same, their optimizers would also be the same. Thus, it follows that the optimal policy for DID
model as stated in Theorem~\ref{lem:opt1}, is also optimal for the CID
model.  This is counter-intuitive, since we expect that, because
of the two-fold trade-off in the former case, the asymptotic value
would be lesser than in the latter. An explicit proof is given in Appendix \ref{proof:lem:fixpopvt}.
 
\begin{figure}
      
  \begin{center}
 
    \begin{subfigure}{\columnwidth}
    \begin{center}
    \includegraphics[width=3.40 in]{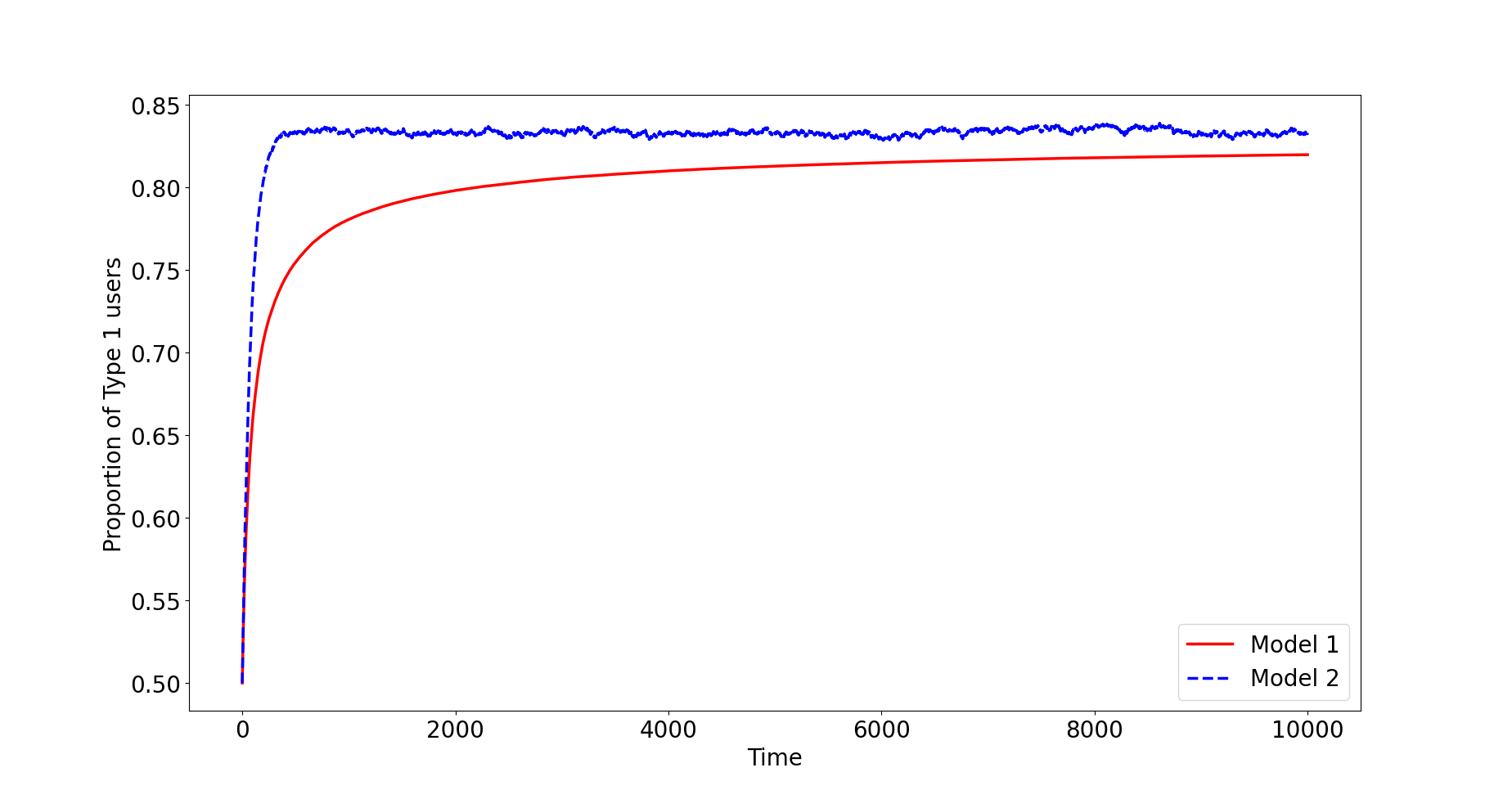}

    \caption{$B_1 = (b_{11}=0.7, b_{12}=0.1, b_{21}=0.2, b_{22}=0.5).$
    Optimal policy is $(p=0,q=0).$}
    \end{center}
    \end{subfigure}

    \begin{subfigure}{\columnwidth}
    \begin{center}
    
     \includegraphics[width=3.40in]{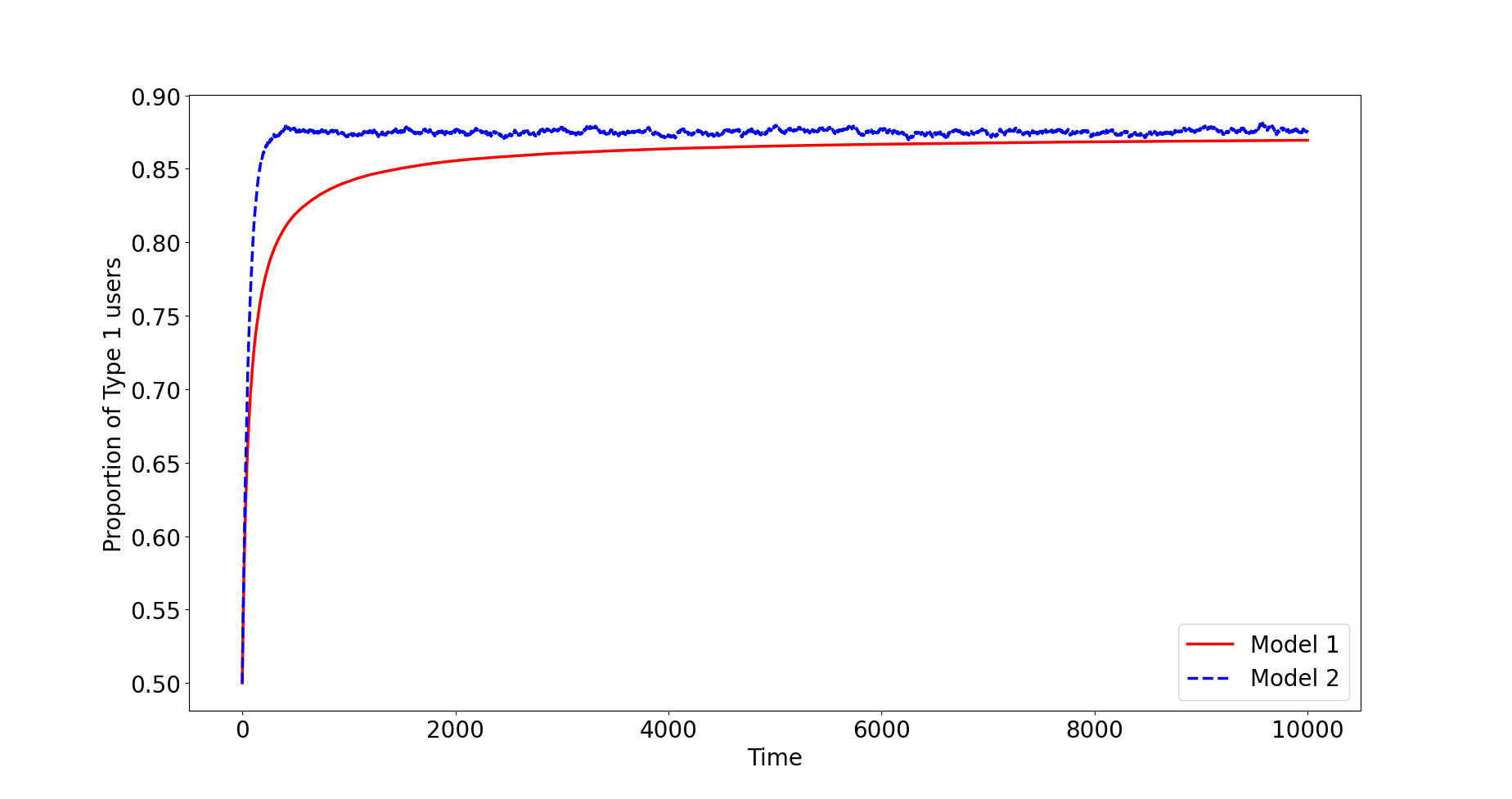}

    \caption{$B_2 = (b_{11}=0.9, b_{12}=0.7, b_{21}=0.7, b_{22}=0.9).$
    Optimal policy is $(p=1,q=0).$}
    \end{center}
    \end{subfigure}
    
  \end{center}

  \caption{Expected population proportion vs time for optimal policies
  that knows $B.$ for DID model and the CID model. The $B$ used for each
  of the plots is also shown.}  \label{fig:fixpopvtime}
\end{figure}

A comparison between the trajectories of $z(t)$ for the optimal
$(p,q)$ for both the decreasing and the constant influence models is
shown in Figure~\ref{fig:fixpopvtime} for some sample values. We see
that in the CID model, the asymptotic value is reached much more
quickly due to the exponential decay term in \eqref{eq:CI-z1-of-t} in
Theorem~\ref{lem:fixpopvt} as opposed to polynomial
in \eqref{eq:DI-z1-of-t} for the DID model.

Next, we show that the other results, i.e., the bounds on the one-step
regret, the cumulative regret for the special case of $b_{11}=b_{22}$
and $b_{12}=b_{21},$ and the cumulative regret for the Thompson
sampling scheme of Algorithm~\ref{algo:TS}, also carry over from the
DID model to the CID model.

\begin{lem}
  \label{lem:fixpopreg}
  Let $\Delta_1:= |b_{11} + b_{12} - 1|$ and $\Delta_2:= |b_{22} +
  b_{21} - 1|$ and let $(p_t,q_t)$ be the strategy in slot $t.$  The
  general expression for regret for the CID model is
  \begin{align*}
    R_t(p_t,q_t) &= z_1(t)|\mathbbm{1}_{\{ b_{11} + b_{12} - 1 > 0\}} -
    p_t|\Delta_1+\\ &\qquad (1-z_1(t))|\mathbbm{1}_{\{ b_{21} + b_{22} - 1 <
      0\}} - q_t|\Delta_2
  \end{align*}
\end{lem}
%\DM{Seems that we do not need $\pi$ in the superscript. How about $R_t(p, q)$ or $R_t(p,q)$?}
Note that the general expression for regret for the model presented in
this section is same as the expression we obtained for the previous
model in Lemma~\ref{lem:regexp}. Thus it follows from the above result that
all regret bounds stated in Section~\ref{sec:unknown-B} hold for the
constant influence model.

\begin{thm}
  \begin{enumerate}

  \item For the special case of $b_{11}=b_{22}$ and $b_{12}=b_{22},$
  the upper bound on the cumulative regret is given
  by \eqref{eq:etcBound1}.

  \item The cumulative regret bound for the ETC algorithm
  (Algorithm~\ref{algo:ETC}) with population dynamics following CID
  model is given by \eqref{eq:etcBound2}.
 
  \item The cumulative regret bound for the Thompson sampling algorithm
  (Algorithm~\ref{algo:TS}) with population dynamics following CID model
  is given by \eqref{eq:tsBound1}.
  \end{enumerate}
\end{thm}

%\DM{This ends abruptly; Could we say a bit more? Probably some plots
%to compare the two Models?}

\DM{Do we need to close this section with some remarks and/or numerical results}

\section{Extensions}
\label{sec:extns}

We consider two different extensions in this section.

\subsection{Generalizing to $N$  arms}
\label{sec:general}

We now consider an MAB with $N$ arms and $N$ user types. As with the
two-type case, the evolution of the population types is tracked using
an urn that has balls of $N$ different colours. Following from the
previous sections, $B[[b_{ij}]]$ is an $N \times N$ matrix of reward
means, where $b_{ij}$ is the mean reward obtained when a user of type
$i$ is shown arm $j.$ The mechanism for user arrival is the same as
that in the two-arm case: Denoting the fraction of balls of color $i$
in the urn at time $t$ by $z_i(t),$ the user is of type of $i$ with
probability $z_i(t).$ Furthermore, we consider the contextual bandit
in which $S$ knows the type of the user.

We consider the following natural extension to the decreasing
influence dynamics model for the two-arm case of
Section~\ref{sec:model}.
\begin{itemize}
\item If $A_t \neq X_t$ then $Z_{A_t}(t+1) = Z_{A_t}(t) + W_t$ and
  $Z_{X_t}(t+1) = Z_{X_t}(t) + 1 - W_t$. This update is exactly like
  the updates of the two-arm DID model of Section~\ref{sec:model}.
\item If $A_t = X_t$ then
  \begin{itemize}
  \item If $W_t = 1$ then $Z_{A_t}(t+1) = Z_{A_t}(t) + 1$. This is 
  also like in Section~\ref{sec:model}. 
  \item If $W_t = 0$ then $Z_{j}(t+1) = Z_{j}(t) + 1$ where $j (\neq
    A_t)$ is an arm chosen uniformly at random from the set of all
    arms excluding arm $A_t$. Note that for $N=2,$ this reduces to
    the two-arm DID model of Section~\ref{sec:model}. 
  \end{itemize}
\end{itemize}
%where $Z_{(\cdot)}(t)$ is the number of balls of a specific type.

In each time step $t$, the MAB uses a randomised policy defined by an
$N \times N$ stochastic matrix $P=[[p_{ij}]],$ where $p_{ij}$ is the
probability that arm $A_j$ is shown to a user of type $i.$
%$\sum_{j} P_{ij} = 1 \forall i$.
As before, we will consider a contextual bandit where $S$ knows the
type of the user before recommending. Further, we will also assume
that $S$ knows the population profile determined by the number of
balls of the different colors in the urn, i.e., $S$ knowns $Z(t) =
[Z_1(t),Z_2(t),\ldots,Z_N(t)]^T$ for all $t.$ As before, we seek the
optimal policy $P^{*}$ for maximizing $E[\Delta Z_1(t)|Z(t)].$ The
following lemma gives us the optimal policy, the extension to
Theorem~\ref{lem:opt1}.

\begin{lem}\label{lem:optNarm}
  The optimal policy $P^{*}$ for the $N$-arm decreasing influence
  dynamics model is given by  
  \begin{itemize}   
  \item For Row 1 of $P^{*}$ :
    \begin{itemize}
    \item $P^{*}_{11} = 1$ if $B_{11} > max(1-B_{12},\ldots,1-B_{1N}) $
    \item else $P^{*}_{1j} = 1$ where $j = \arg \max_k(1-B_{1k})$
    \end{itemize}
  \item For Row $i$ ($\neq 1$) of $P^{*}$ :
    \begin{itemize}
    \item $P^{*}_{i1} = 1$ if $B_{i1} > (1-B_{ii})/(N-1)$
    \item $P^{*}_{ii} = 1$ otherwise
    \end{itemize}
    
  \end{itemize}
\end{lem}

We remark here that this policy is more generally applicable. For
example, even if the population dynamics were changed as follows: for
the case $A_t = X_t$ and $W_t=0,$ then choose $j$ with probability
proportional to the $j$-type population instead of choosing uniformly
at random, the optimal policy would be of a form similar to that in
Lemma~\ref{lem:optNarm}. This is because the derivation would still
involve maximizing a convex combination like the one we see while
proving the lemma (see the Appendix).

For the case when the matrix $B$ is unknown, once we have the
expression for an optimal policy for the $N$-arm model (like the one
in Lemma \ref{lem:optNarm}), we can apply Thompson sampling (with the
optimal policy applied on a sampled matrix $\Tilde{B}$ in each time
slot instead of $B$) to maximize the proportion of type 1 users.

\begin{figure}
      
  \begin{center}
    \begin{subfigure}{\columnwidth}
    \begin{center}
    \includegraphics[width=3.40 in]{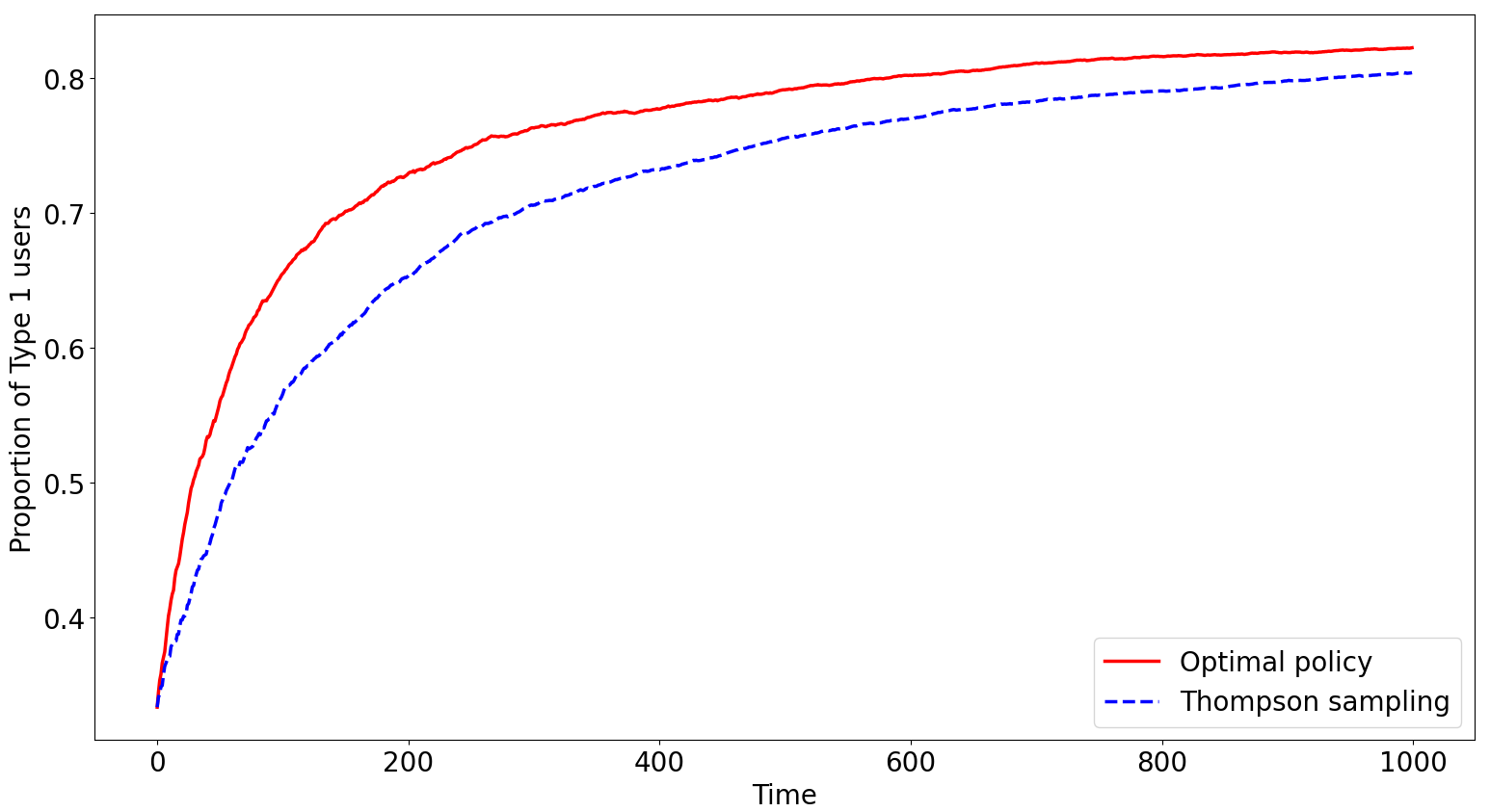}

    \caption{Preference shaping with 3 arms. Matrix $B$ is $3 \times
    3$ with diagonal terms = 0.9 and off-diagonal terms =
    0.7.}
    \end{center}
    \end{subfigure}

    \begin{subfigure}{\columnwidth}
    \begin{center}
    
     \includegraphics[width=3.40in]{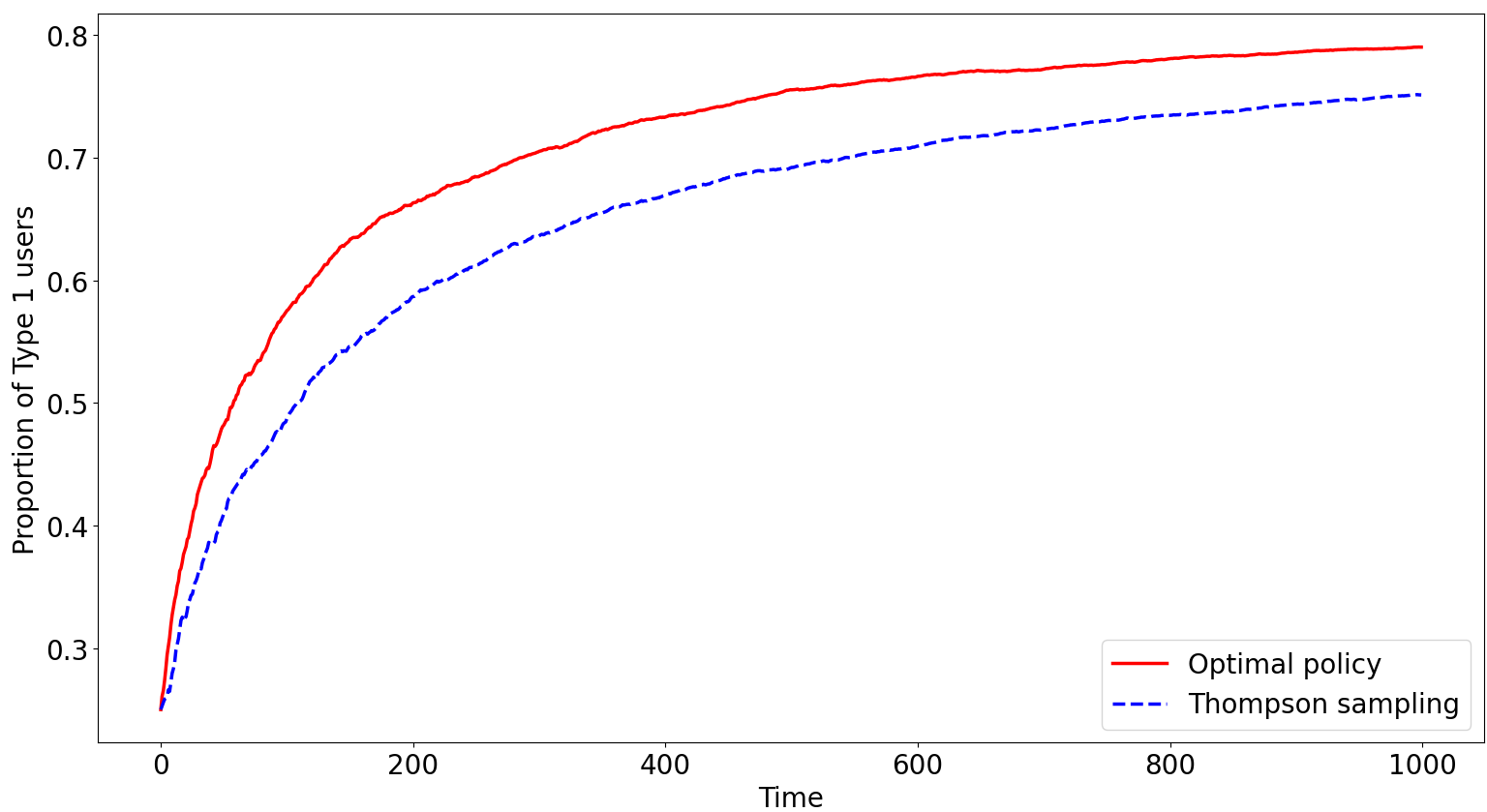}

    \caption{Preference shaping with 4 arms. Matrix $B$ is $4 \times
    4$ with diagonal terms = 0.9 and off-diagonal terms = 0.6.}
    \end{center}
    \end{subfigure}
    
    \begin{subfigure}{\columnwidth}
    \begin{center}
    
     \includegraphics[width=3.40in]{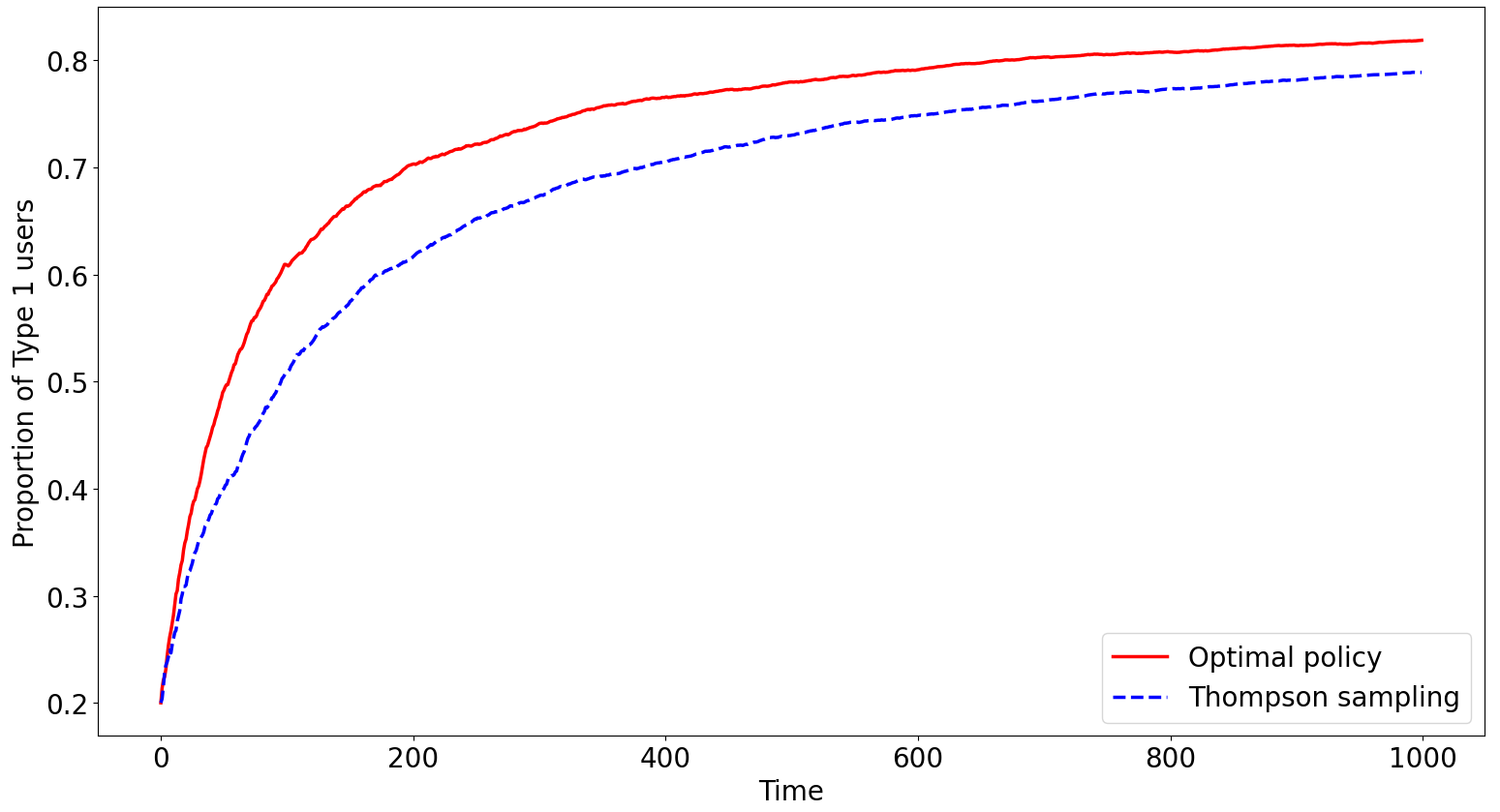}

    \caption{Preference shaping with 5 arms. Matrix $B$ is $5
      \times 5$ with diagonal terms = 0.9 and off-diagonal terms =
                        0.7.}
    \end{center}
    \end{subfigure}

  \end{center}

  \caption{ Expected population proportion vs time for optimal policy
    and Thompson sampling policy for $N-$arm case. The $N$ and $B$
    used for each of the plots is also
    shown. } \label{fig:popvtimeNarms}
\end{figure}

The optimal and Thompson sampling based strategies mentioned above
have been simulated for the $N-$arms case and the trajectory of the
type 1 population for a few example cases are shown in
Figure~\ref{fig:popvtimeNarms}. Even in the $N>2-$armed bandit
examples, we observe that the Thompson sampling does not fall far
behind the optimal trajectory and gives us a healthy majority of type
1 users ($\approx 80\%$ in all cases). We have not obtained analytic
guarantees on the population trajectories for these cases, and a
regret-based analysis remains open.

\subsection{Two competing recommendation systems}
\label{sec:twoSys}

Consider the same two-arm model as before, but we have two
recommendation systems $S_1$ and $S_2$ instead of just one. Assume
that $S_1$ is trying to maximize the number of type 1 users while
$S_2$ is trying to achieve the opposite. Had the recommendation
systems been alone, we saw that they may adopt optimal policies that
recommend a disliked arm to a user. In real life, such an action might
make the user dislike the recommendation system itself. Therefore,
when multiple systems are present, we need to keep track of a
``popularity'' metric that measures how popular a particular system is
among a certain type of user. In a time slot, a user either goes to be
served by $S_1$ or $S_2$. The probability of a user going to a
recommendation system is determined by the \textit{popularity} of the
system among users of the same type. The popularity of a system should
increase on receiving a positive reward and decrease on receiving a
negative (or zero) reward. We now make this more precise.

\textbf{Competing recommendation systems (CRS):} 
Define the popularity matrix $P^t = \begin{bmatrix} P_{11}^t &
P_{12}^t \\ P_{21}^t & P_{22}^t
\end{bmatrix}$.

The row number of an element represents the user type, and the column
number represents the recommendation system. If a user of type $X_t$
arrives, it chooses to be served by the system $S_1$ with probability
$\frac{P_{X_t1}^t}{P_{X_t1}^t + P_{X_t2}^t}$. For the case when the
user goes with system $S_1$, we update the popularity matrix in the
following way. 
\begin{align}
  P_{X_t1}^{t+1} &= P_{X_t1}^{t} + W_t\\
  P_{X_t2}^{t+1} &= P_{X_t2}^{t} + (1-W_t)
\end{align}
where $W_t$ is the reward obtained after an arm is recommended by the
system. This is similar to the population preference dynamics. The
population of users is updated in the DID or the CID models from
Section~\ref{sec:model}. Thus we have both the population preference
for arm and the population preference for the recommendation system
interacting with the recommendations made and the rewards seen by the
users.

In this setup of recommendation systems with opposing objectives, an
interesting question is whether there are any equilibrium strategies
that balance the popularity of the recommendation system, akin to
market share, as well as their population preference shaping
objectives. If the objective of $S_1$ and $S_2$ were to only increase
their popularity, they would both follow the policy
$(p=1,q=1)$. However, since the optimal population preference shaping
policy might not always be $(p=1,q=1)$, each faces a tradeoff between
their goals of increasing popularity and of preference shaping. In
this preliminary study, we assume that each system is concentrating
solely on preference shaping and ran some simulations. We note that
depending on the structure of matrix $B$, we obtain two kinds of
behaviours.

\subsubsection{\textbf{Case 1 (Uniform Population)}}
This is the case where the matrix $B$ is such that the optimal policy
in Theorem~\ref{lem:opt1} is either $(0,0)$ or $(1,1)$. Note that the
optimal policy of $S_1$ and $S_2$ come out to be exactly the opposite
of each other (i.e. if $(p^*,q^*) = (1,1)$ for $S_1$, then it is
$(0,0)$ for $S_2$ and vice-versa).

\begin{figure}
  \begin{center}
    \includegraphics[width=3.4in]{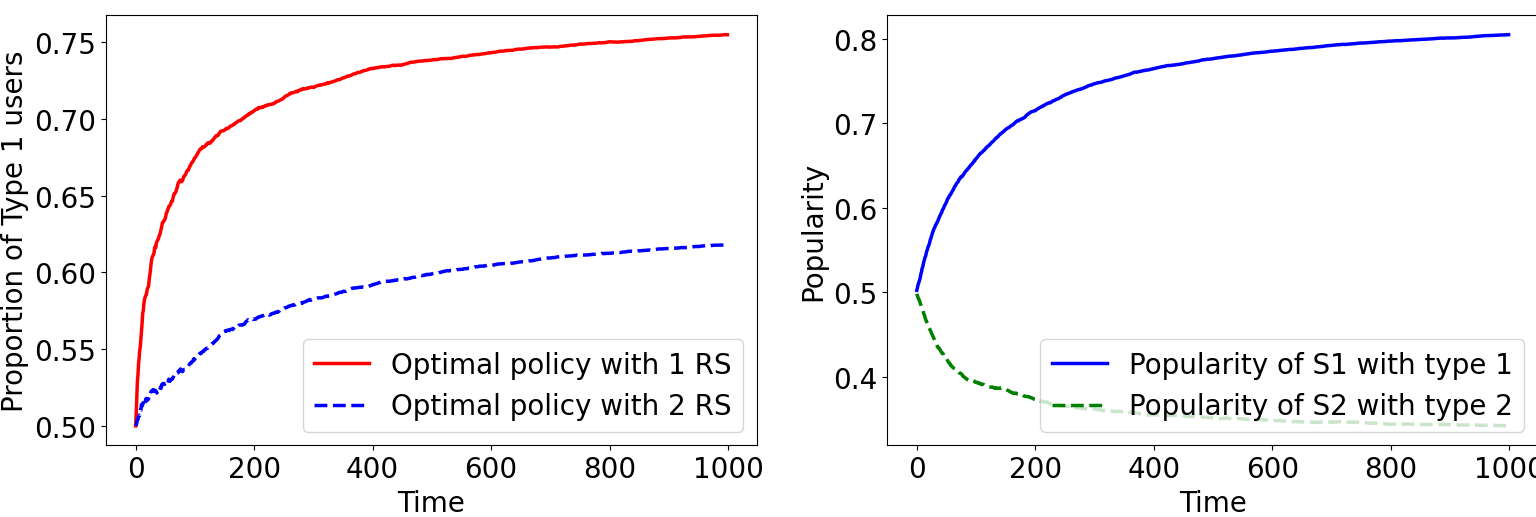}
  \end{center}

  \caption{Expected population proportion vs Time (left) and
  Popularity vs Time (right) for optimal policies for $B_1 =
  (b_{11}=0.9, b_{12}=0.4, b_{21}=0.2,
  b_{22}=0.6).$}\label{popvtime21}
  
\end{figure}

\begin{figure}
  \begin{center}
    \includegraphics[width=3.4in]{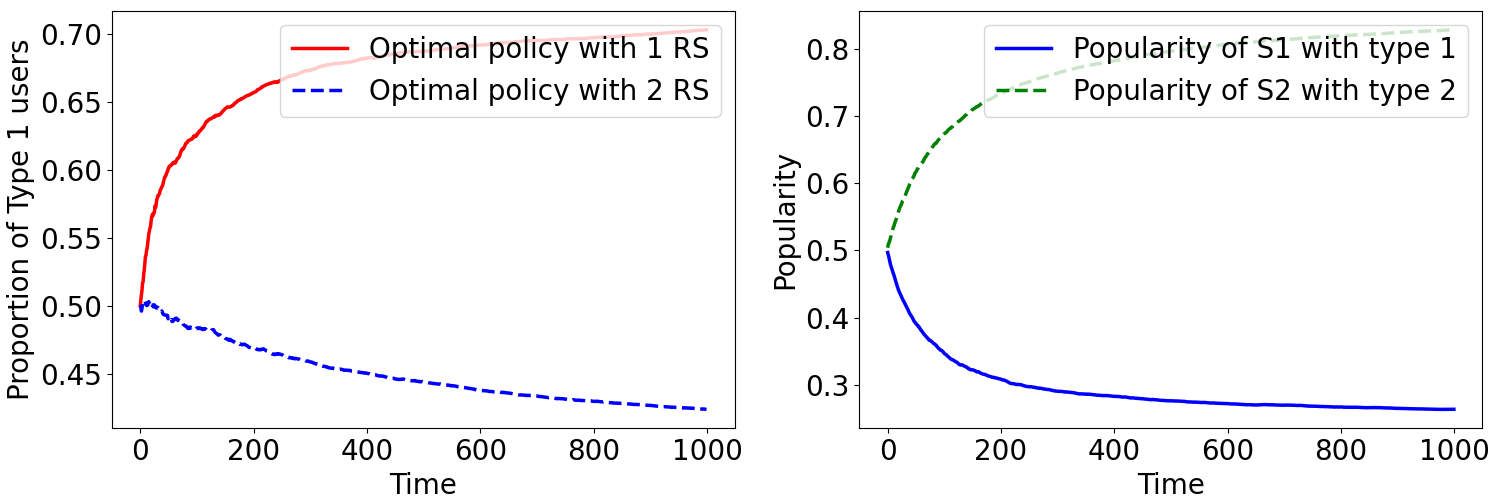}
  \end{center}

  \caption{Expected population proportion vs Time (left) and
  Popularity vs Time (right) for optimal policies for $B_4 =
  (b_{11}=0.7, b_{12}=0.1, b_{21}=0.3,
  b_{22}=0.9).$} \label{popvtime24}
\end{figure}

In this case, the popularity of the same recommender system dominates
in both type of user populations; see Fig.~\ref{popvtime21}
and \ref{popvtime24}). In the former case, the optimal policy for
$S_1$ is $(1,1)$. Here, $S_1$ clearly has no incentive to deviate from
this policy since this policy increases both type 1 preferences and
its own popularity among all users. At the end of 1000 time slots,
therefore, we observe that a large fraction of the user population is
using the services of $S_1.$ This makes the type 1 users the majority
in the population. Note that this majority is still less than what it
could have been had $S_2$ not been present. In Fig.~\ref{popvtime24},
the opposite happens, i.e., $S_2$ dominates.

In both cases above, a large fraction of the population end up
preferring one recommender over the other, regardless of their type.

\subsubsection{\textbf{Case 2 (Polarized Population)}}
This is the case where the matrix $B$ is such that the optimal policy
in Theorem~\ref{lem:opt1} is either $(0,1)$ or $(1,0)$.

\begin{figure}
  \begin{center}
    \includegraphics[width=3.4in]{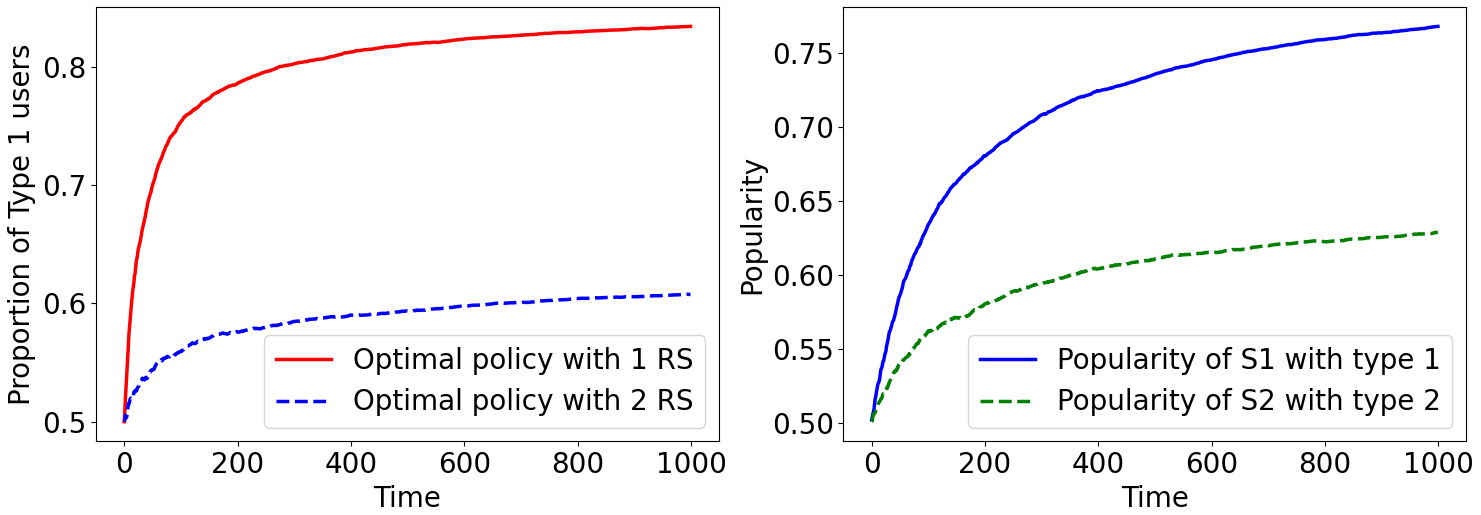}
  \end{center}

  \caption{Expected population proportion vs Time (left) and
  Popularity vs Time (right) for optimal policies for $B_2 =
  (b_{11}=0.7, b_{12}=0.5, b_{21}=0.6, b_{22}=0.8).$
  }\label{popvtime22}
\end{figure}

\begin{figure}
  \begin{center}
    \includegraphics[width=3.4in]{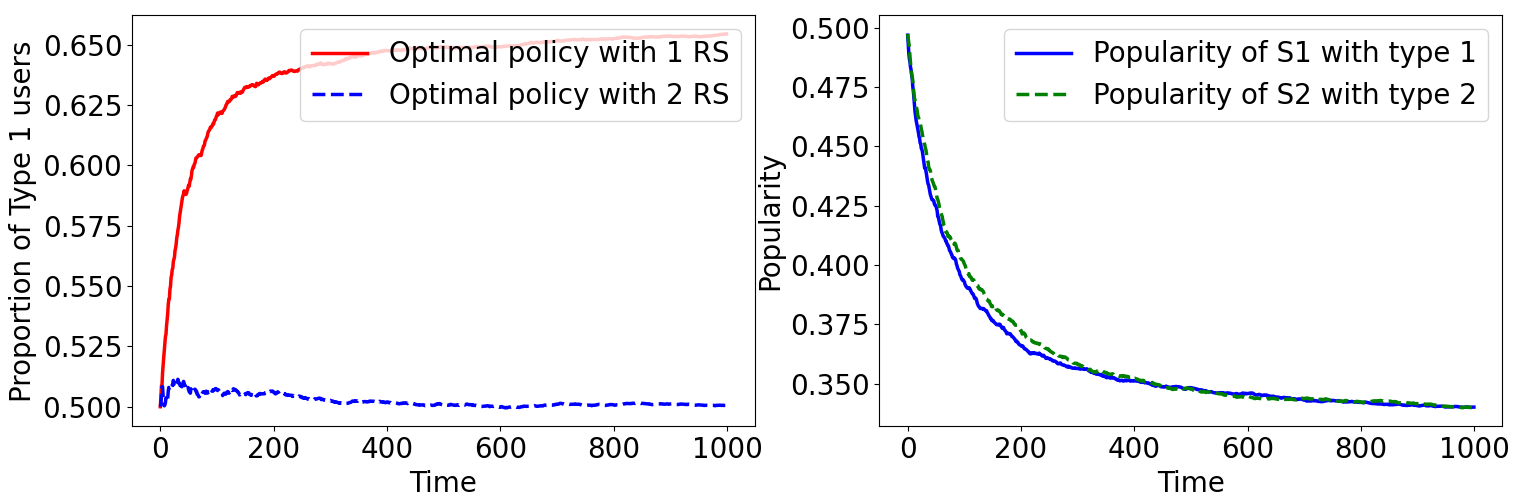}
  \end{center}

  \caption{Expected population proportion vs Time (left) and
  Popularity vs Time (right) for optimal policies for $B_3 =
  (b_{11}=0.6, b_{12}=0.2, b_{21}=0.2,
  b_{22}=0.6).$} \label{popvtime23}
\end{figure}

In this case, the popularity of the different recommender system
dominates in different type of user populations (See
Fig.~\ref{popvtime22} and \ref{popvtime23}). Take the former case
where the optimal policy for $S_1$ is $(1,0)$. At the end of 1000 time
slots, we observe that the vast majority of the user population of
type 1 is using the services of $S_1$, and those of type 2, are using
the services of $S_2$. In the next figure, the opposite happens.

In both cases, we see that each type of user dominates the user base
of different recommendation systems.

%\section{Related Literature}
%\label{sec:literature}

%\input{lit-survey}

\section{Concluding Remarks}
\label{sec:conclude}
In this paper, we considered the problem of
  preference shaping in a user population using multi-armed bandit
  algorithms. We presented a simplified version of this problem for
  two arms and two corresponding user types, where the preferences of
  a user change in response to the random Bernoulli reward obtained
  (can be thought of as like/dislike) in response to the
  recommendation. We found how the population behaves if the
  recommender follows a certain policy and hence found the optimal
  policy to be followed if the Bernoulli reward means are known. We
  then used Explore-then-Commit and Thompson sampling algorithms to
  get policies that approach the optimal in the case when the
  Bernoulli reward means are unknown. We also extended these
  algorithms for the cases when we have $N$ arms or more than one
  recommender.

Clearly both the DID and the CID models can be generalised in many
ways. For example, in the DID model, the composition of the urn could
be changed by adding $K$ balls to the urn at every step of which $m$
balls are of one color and $K-m$ balls are of the other color. The
rate at which the influence decreases can be tuned using different $K$
and $m.$ The theory for this would be similar to that derived in this
paper. Similarly, in the CID model, the composition of the urn could
be changed with probability $\beta$ in every step. Once again, a
suitable choice of $\beta$ will determine the degree of change in
every step. 

There are two possible directions in which one can further extend our
models and results. Firstly, we can generalize by evolving the type of
a user via an underlying Markov Decision Process rather than the urn
model that we considered.  Note that mapping our model to an MDP
requires an MDP \textit{where the transition probabilities are
  affected by the rewards that are accrued}. Since many recommendation
systems may not have an estimate about the composition of the user
population, one can further extend this to the case where the type of
a user is not visible.

A second direction would be to keep the existing two recommendation
system model and introduce different rewards for opinion shaping and
popularity. This maps to the problem that a real-life recommendation
system may have between, say PR/advertising money (opinion shaping)
and maintaining popularity among its current user base. This model
would provide a trade-off between the algorithms discussed in
\cite{Shah18} and the ones discussed in this paper, and it would be
interesting to see whether we arrive at some equilibrium strategy and
population distribution for such a setting.

Understanding and modeling the interaction between a recommender
system's learning algorithm and the preferences of the user population
is becoming increasingly important. Specifically, the `exploration'
part of the algorithm could `expose' the user to possibilities that in
turn might make them to also explore and possibly start preferring
other options. Similarly, the exploitation part of the algorithm may
reinforce the users's possibly weak preferences. We believe that the
models that we have presented here could be used in understanding and
modeling such behaviour.

\bibliographystyle{IEEEtran}
\bibliography{../../../Multiarmed-Bandits.bib,../../cit.bib}

\appendix

\subsection{Comparison between regret definitions}\label{sec:regComp}
\DM{An optimality notion could be to maximize $\sum_{i}^T Z_1(t).$
  Seems to me that what is being done below is not this. Should we
  discuss this? }

The commonly used definition of regret would compare the difference in
the population trajectories of the optimal and the candidate policies
directly. Specifically, let $Z_1^{*}(t)$ be the trajectory of the
population of type 1 balls in the urn when the optimal policy is
applied and let $Z_1(t)$ be the population when the candidate policy
is applied. Thus the following definition of regret, denoted by
$R_t^{'},$ is more like what is commonly used in the literature when
the objective is to learn the best arm.
\begin{equation}
  R_t^{'} = E[\Delta Z_1^{*}(t)|Z_1^{*}(t)] - E[\Delta Z_1(t)|Z_1(t)]
  \label{eq:std-regret}
\end{equation}
Let us now compare this with our definition of regret $R_t$ in
\eqref{eq:regret}. Taking the difference between the two definitions
gives us
\begin{align*}
  R_t - R_t^{'} &= E[\Delta Z_1^{*}(t)|Z_1(t)] - E[\Delta
  Z_1^{*}(t)|Z_1^{*}(t)]\\
  &= (z_1(t) - z_1^{*}(t)) \left( (b_{11} + b_{12} - 1)^+ + (1-b_{12})  
   \right. \\
  &\hspace{0.5in} \left. - (1-b_{21}-b_{22})^+ - b_{21} \right),  
\end{align*}
where $x^+ = \max(x,0).$ In the second equality above we have used the
optimal policy for the DID model obtained in Theorem~\ref{lem:opt1} to
derive the expressions for $E[\Delta Z_1(t)|Z_1(t)]$.  (From the
results in Section~\ref{sec:voter} this is also applicable to CID
model.) We know, by definition, that the trajectory followed by the
optimal policy is always above that followed by any other
policy. Therefore, $(z_1(t) - z_1^{*}(t)) < 0$. Thus the lemma below
follows immediately.
\begin{lem}
  \label{lem:regComp}
  $R_t^{'}$ is bounded above by the regret $R_t$ if and only if $(b_{11} + b_{12}
  - 1)^+ + (1-b_{12}) \leq (1-b_{21}-b_{22})^+ + b_{21}$.
\end{lem} 

\DM{Not sure of the above proof; see remarks on the definition of
  optimality above. Also need to say a bit more about the what this
  lemma tells us. }
%In the rest of the paper, we will use the definition $R_t$. 

\subsection{Proof of Theorem~\ref{lem:opt1}}
\label{sec:lem:opt1}

The expression for expected increase in type 1 balls in time slot $t$
is given by
\begin{align*}
  E\left[ \Delta Z_1^{\pi}(t)|z_1(t) \right]
  &=  P(\Delta Z_1^{\pi}(t) = 1|z_1(t))\\
  &= \sum_{i = 1,2} P(X_t = i|z_1(t))P(\Delta Z_1^{\pi}(t) = 1, \\
  &	\qquad	| X_t = i,z_1(t))\\
  &= z_1(t)(p_t(b_{11}) + (1-p_t)(1-b_{12})) + \\& \qquad(1
  -z_1(t))((1-q_t)(b_{21}) + (q_t)(1-b_{22}))\\
  &= z_1(t)(p_t(b_{11} + b_{12} - 1) + 1-b_{12}) +  \\
  & \qquad (1 -z_1(t))(q_t(1-b_{22}-b_{21}) + b_{21})
\end{align*}
The last expression above is to be maximized over all possible
$(p_t,q_t)$. This is a simple linear expression in terms of both these
variables. Hence, since $z_1(t),1-z_1(t) > 0$, $E[\Delta
  Z_1^{\pi}(t)|z_1(t)]$ is maximized when
$$
(p_t,q_t) = (\mathbbm{1}_{\{ b_{11} + b_{12} - 1 >
  0\}},\mathbbm{1}_{\{1- b_{21} + b_{22} > 0\}}). 
$$
which proves the theorem 
\qed

\subsection{Proof of Lemma~\ref{lem:prop1}}
\label{sec:lem:prop1}

For $d_1$ and $d_2$ as defined in the statement of the lemma, and
referring to the proof of Lemma~\ref{lem:opt1}, we get:
\begin{align*}
  E[\Delta Z_1^{\pi}(t)|z_1(t)] &= z_1(t)(1-d_1) + (1 - z_1(t))d_2
\end{align*}
This corresponds to o.d.e
\begin{displaymath}
  \Dot{Z_1^{\pi}}(t) = \frac{Z_1(t)(1-d_1) + (N_0 + t -
    Z_1(t))d_2}{N_0 + t}
\end{displaymath}
We can now substitute $Z_1^{\pi}(t) = (N_0 + t)z_1(t)$ to get the
o.d.e.
\begin{displaymath}
  \Dot{z_1}(t) = \frac{d_2 - (d_1 + d_2)z_1(t)}{N_0 + t}
\end{displaymath}
Solving this o.d.e gives us the desired result. \qed

\subsection{Proof of Theorem~\ref{thm:propmax}}
\label{sec:thm:propmax}

The proportion $\frac{d_2}{d_1 + d_2}$ is to be maximised over the
variables $(p,q)$. Since $d_1$ is only a function of $p$, we first
minimize that with respect to $p$ since it is in the denominator.
\begin{align*}
  p^{*} &= \arg \min_p \ d_1 \\
  &= \arg \min_p \ p(1-b_{11}) + (1-p)b_{12}\\
  &= \arg \min_p \ p(1-b_{11}-b_{12}) + b_{12}\\
  &= \mathbbm{1}_{\{ b_{11}+b_{12}-1 > 0\}}
\end{align*}
Now, $d_2$ (which depends only on $q$) appears both in the numerator
and the denominator. But we can use the easily verifiable fact that
$\frac{d_2}{d_1 + d_2}$ is a strictly increasing function of
$d_2$. This means maximizing $d_2$ also maximizes the proportion.
\begin{align*}
  q^{*} &= \arg \max_q \ d_2 \\
  &= \arg \max_q \ q(1-b_{22}) + (1-q)b_{21}\\
  &= \arg \max_q \ q(1-b_{22}-b_{21}) + b_{21}\\
  &= \mathbbm{1}_{\{ b_{22}+b_{21}-1 < 0\}}
\end{align*}
We see that $(p^{*},q^{*})$ derived here indeed match with the
results in Lemma~\ref{lem:opt1}

\qed
\subsection{Proof of Lemma \ref{lem:regexp}}
\label{sec:lem:regexp}

We know that:
\begin{align*}
  \Delta Z^{\pi}_1(t) &= W_t \mathbbm{1}_{A_t = 1} + (1-W_t)
  \mathbbm{1}_{A_t = 2} \\
  &= \Tilde{b}_{11} \mathbbm{1}_{A_t = 1, X_t = 1} + \Tilde{b}_{21}
  \mathbbm{1}_{A_t = 1, X_t = 2} + \\
  & \qquad (1-\Tilde{b}_{12}) \mathbbm{1}_{A_t = 2, X_t = 1} +
  (1-\Tilde{b}_{22}) \mathbbm{1}_{A_t = 2, X_t = 2}
\end{align*}
where $\Tilde{b}_{ij}$ are independent samples of the Bernoulli
rewards with the respective means as $b_{ij}$.

Therefore,
\begin{align*}
  E[\Delta Z^{\pi}_1(t)|z_1(t)] &= b_{11}z_1(t)p_t + b_{21}(1 -
  z_1(t))(1 - q_t)+ \\
  & \qquad (1-b_{22})(1-z_1(t))q_t+\\
  & \qquad (1-b_{21})z_1(t)(1-p_t)\\
  &= z_1(t)p_t(b_{11} + b_{12} - 1)+ \\
  & \qquad (1-z_1(t))q_t(1-b_{22}-b_{21})+  \\
  &  \qquad b_{21}(1-z_1(t)) + (1-b_{21})z_1(t)
\end{align*}
For the optimal policy, the same expression becomes:
\begin{align*}
  E[\Delta Z^{*}_1(t)|z_1(t)] &= z_1(t)\mathbbm{1}_{\{ b_{11} + b_{12}
    - 1 > 0\}}(b_{11} + b_{12} - 1)+ \\
  & \qquad (1-z_1(t))\mathbbm{1}_{\{ b_{21} + b_{22} - 1 <
    0\}}(1-b_{22}-b_{21})+ \\
  & \qquad b_{21}(1-z_1(t)) + (1-b_{21})z_1(t)
\end{align*}

Substituting these results in the definition \ref{def:regret}, we get :
\begin{align*}
  R_{t}^{\pi} &= E[\Delta Z_1^{*}(t) - \Delta Z_1^{\pi}(t)|Z_1^{*}(t)
    = Z_1^{\pi}(t)]\\
  &=z_1(t)(\mathbbm{1}_{\{ b_{11} + b_{12} - 1 > 0\}} - p_t)(b_{11} +
  b_{12} - 1)+  \\
  & \qquad (1-z_1(t))(\mathbbm{1}_{\{ b_{21} + b_{22} - 1 < 0\}} -
  q_t)(1-b_{21} - b_{22})
\end{align*}
Since $0 \leq p_t,q_t \leq 1$, it means that $(\mathbbm{1}_{\{ b_{11}
  + b_{12} - 1 > 0\}} - p_t)(b_{11} + b_{12} - 1)$ and
$(\mathbbm{1}_{\{ b_{21} + b_{22} - 1 < 0\}} - q_t)(1-b_{21} -
b_{22})$ are always positive. This proves the desired result.
\qed
\subsection{Proof of Lemma \ref{lem:regexpetc}}
\label{sec:lem:regexpetc}

The expression for $R_{explore}$ is obtained by substituting
$(p_t,q_t) = (0.5,0.5)$ in the regret formula of Lemma \ref{lem:regexp}
and summing from $t = 1$ to $t = m$.

The expression for $R_{commit}$ is obtained by substituting $(p_t,q_t)
= (\mathbbm{1}_{\{ \hat{b}_{11} + \hat{b}_{12} - 1 >
  0\}},\mathbbm{1}_{\{ \hat{b}_{22} + \hat{b}_{21} - 1 < 0\}})$ in the
regret formula of lemma \ref{lem:regexp} and summing from $t = m$ to $t =
T$.
\qed
\subsection{Proof of Theorem~\ref{ETC}}

For $b_{11} = b_{22}$ and $b_{12} = b_{21}$, we get
\begin{align*}
  R_{explore} &= \frac{m\Delta_1}{2}\\
  R_{commit} &= p_{err}(T-m)\Delta_1
\end{align*}
The term $p_{err}$ can be bounded in the following way:
\begin{align}
  p_{err} &= P((\hat{b}_{11} + \hat{b}_{12} - 1)(b_{11} + b_{12} - 1) < 0) \label{subset}\\
  &\leq P(|\hat{b}_{11} + \hat{b}_{12} - b_{11} - b_{12}| > \Delta_1) \label{superset}
\end{align}
This is because the event in \ref{subset} is a subset of the event in
\ref{superset}.

Now, we know that Bernoulli random variables are
$1/2-$sub-Gaussian. Since we show arm 1 and 2 with equal probability in
the exploration phases, the expected number of times $b_{11}$ and
$b_{12}$ are sampled is $m/2$ for each of the arms. This implies that
$\hat{b}_{11} - b_{11}$ is $1/\sqrt{m/2}-$sub-Gaussian, which in turn
implies that $\hat{b}_{11} + \hat{b}_{12} - b_{11} - b_{12}$ is
$2/\sqrt{m}-$sub-Gaussian. Therefore we can use the Hoeffding bounds on
sub-Gaussian random variables to further put a more useful bound on
$p_{err},$ i.e.,
\begin{displaymath}
  p_{err} \leq e^{-m\Delta_1^2/8}. 
\end{displaymath}

Therefore, we get a $\Delta_{i}$ dependent bound on the regret to be 
\begin{displaymath}
  R_{explore} + R_{commit} \leq m\Delta_1/2 +
  (T-m)\Delta_1e^{-m\Delta_1^2/8}. 
\end{displaymath}
\qed
\subsection{Proof of Theorem~\ref{thm:Thompson1} }

Using Lemma \ref{lem:regexp}, the expression for the regret $R_t$ in time
slot $t$ for Thompson sampling becomes:
\begin{displaymath}
  R_t = z_1(t)p^{err}_t\Delta_1 + (1-z_1(t))q^{err}_t\Delta_2
\end{displaymath}
where
\begin{align*}
  p^{err}_t & = P((\Tilde{b}_{11}^t + \Tilde{b}_{12}^t - 1)(b_{11} +
  b_{12} - 1) < 0) \\
  q^{err}_t & = P((\Tilde{b}_{22}^t + \Tilde{b}_{21}^t - 1)(b_{22} +
  b_{21} - 1) < 0)
\end{align*}

Here, the variables $\Tilde{b}_{ij}^t$ denote the sampled matrix
elements based on the $Beta(\alpha_{ij},\beta_{ij})$ distribution at
that time.

Let us now put a bound on $p^{err}_t$.
\begin{align*}
  p^{err}_t &\leq P(|\Tilde{b}_{11}^t + \Tilde{b}_{12}^t - b_{11} -
  b_{12}| > \Delta_1)\\
  &\leq \frac{Var(\Tilde{b}_{11}^t + \Tilde{b}_{12}^t)}{\Delta_1^2}
\end{align*}
where the second inequality is a direct application of Chebyschev's
inequality.

Since $\Tilde{b}_{11}^t$ and $\Tilde{b}_{12}^t$ are independently
sampled, we have:
\begin{align*}
  Var(\Tilde{b}_{11}^t + \Tilde{b}_{12}^t) &=
  \frac{\alpha_{11}^t\beta_{11}^t}{(\alpha_{11}^t+\beta_{11}^t)^2(\alpha_{11}^t+\beta_{11}^t
    + 1)} \\
  &+
  \frac{\alpha_{12}^t\beta_{12}^t}{(\alpha_{12}^t+\beta_{12}^t)^2(\alpha_{12}^t+\beta_{12}^t
    + 1)}
\end{align*}
where we have used the expression for the variance of Beta distributed
random variables.
 
Let us assume that $x_{ij}^t=\frac{\alpha_{ij}^t}{\alpha_{ij}^t +
  \beta_{ij}^t} $.  Substituting these in the expression for the
variance, we get,
\begin{align*}
  Var(\Tilde{b}_{11}^t + \Tilde{b}_{12}^t) &=
  \frac{x_{11}^t(1-x_{11}^t)}{\alpha_{11}^t+\beta_{11}^t+1} +
  \frac{x_{12}^t(1-x_{12}^t)}{\alpha_{12}^t+\beta_{12}^t+1}\\
  &\leq \frac{1}{4}\left(\frac{1}{\alpha_{11}^t+\beta_{11}^t+1} +
  \frac{1}{\alpha_{12}^t+\beta_{12}^t+1}\right)\\ &= \frac{1}{4}
  \frac{\alpha_{11}^t+\beta_{11}^t+\alpha_{12}^t+\beta_{12}^t +
    2}{(\alpha_{11}^t+\beta_{11}^t+1)(\alpha_{12}^t+\beta_{12}^t+1)}
\end{align*}
Now, $\alpha_{11}^t+\beta_{11}^t+\alpha_{12}^t+\beta_{12}^t$ is the
expected number of times a type 1 user appears. Out of that, let $f_1$
be the fraction of the time arm 1 was recommended.  Therefore, by
definition, $z^{*}t \geq
\alpha_{11}^t+\beta_{11}^t+\alpha_{12}^t+\beta_{12}^t$ and
$(\alpha_{11}^t+\beta_{11}^t+1)(\alpha_{12}^t+\beta_{12}^t+1) \approx
(f_1 t )((1-f_1)t)$ Substituting these observations, we get:
\begin{align*}
Var(\Tilde{b}_{11}^t + \Tilde{b}_{12}^t) &\leq \frac{z^* t}{4(f_1 t)((1-f_1)t)}\\
&= \frac{z^*}{4f_1(1-f_1)t}
\end{align*}
This gives us:
\begin{displaymath}
  p^{err}_t \leq \frac{z^*}{4\Delta_1^2f_1(1-f_1)t}
\end{displaymath}
Therefore, 
\begin{displaymath}
  R_t \leq \frac{z^*}{t}\left(\frac{z_1(t)}{4\Delta_1f_1(1-f_1)}+\frac{1-z_1(t)}{4\Delta_2f_2(1-f_2)}\right)
\end{displaymath}
where $f_2$ is the fraction of time arm 2 was recommended when a user of type 2 showed up.
By definition, $z^{*} \geq z_1(t),1-z_1(t)$. Therefore, using this and
summing over $t$, we get
\begin{displaymath}
  R_{[1:T]}^{Thomp} \leq
  \frac{(z^{*})^2}{4}\left(\frac{1}{f_1(1-f_1)\Delta_1}+\frac{1}{f_2(1-f_2)\Delta_2}\right)\sum_{t=1}^{T}\frac{1}{t}.
\end{displaymath}
This leads to the desired result by bounding the summation by an
integral.

\qed
\subsection{Proof for Lemma \ref{lem:fixpopreg}}

In the proof of the Lemma \ref{lem:fixpopvt}, we had obtained :
\begin{align*}
E[Z_1(t+1)-Z_1(t)|Z_1(t)] &= p_2 - p_1 \\
\implies E[\Delta Z_1(t)|z_1(t)] &= z_1(t)( p(1-b_{11}) + (1-p)b_{12})+\\&\qquad (1-z_1(t))(q(1-b_{22})+ \\& \qquad (1-q)b_{21})\\
&= z_1(t) p(1-b_{11}-b_{12})+\\&\qquad (1-z_1(t))q(1-b_{21}-b_{22}) +\\
&\qquad z_1(t)b_{12} + (1-z_1(t))b_{21}
\end{align*}
Similarly, we get (since the optimal policy is same as the one given in Lemma \ref{lem:opt1}) :
\begin{align*}
E[\Delta Z^{*}(t)|z_1(t)]& = z_1(t)
p^{*}(1-b_{11}-b_{12})+\\& \qquad (1-z_1(t))q^{*}(1-b_{21}-b_{22})+
\\ & \qquad z_1(t)b_{12} + (1-z_1(t))b_{21}
\end{align*}
where $p^{*} = \mathbbm{1}_{\{b_{11}+b_{12}-1>0\}}$ and $q^{*} =
\mathbbm{1}_{\{b_{22}+b_{21}-1<0\}}$.

Using the expression in definition \ref{def:regret} for regret directly
gives us the desired result.
\qed
\subsection{Proof of Lemma~\ref{lem:fixpopvt}}\label{proof:lem:fixpopvt}

From the model, we know that
\begin{align*}
  Z_1(t+1) &= Z_1(t) - 1 \ \textrm{w.p} \ p_1 = Z_1(t)d_1/N_0\\
		&= Z_1(t) + 1 \  \textrm{w.p} \ p_2 = (1-Z_1(t)/N_0)d_2\\
		&= Z_1(t) \ \textrm{w.p} \ 1-p_1-p_2 \\	  
\end{align*}
Therefore,
\begin{align*}
E[Z_1(t+1)-Z_1(t)|Z_1(t)] &= p_2 - p_1 \\
&= d_2 -\frac{Z_1(t)}{N_0 } \ (d_1 +d_2)\\
\end{align*}
Thus gives us the corresponding o.d.e. as
$$\Dot{Z}_1(t) = d_2 -\frac{Z_1(t)}{N_0 } \ (d_1 +d_2)$$
Solving for
$Z_1$ and dividing the solution by $N_0$ gives us the required
solution for $z_1(t)$.
\qed
\subsection{Proof of Lemma~\ref{lem:optNarm}}\label{sec:nArmproof}
For this model, we have
  \begin{align*}
    E[\Delta Z_1(t)|\Bar{z}(t)] &= P(\Delta Z_1(t) = 1 |\Bar{z}(t) )\\
    &= z_1(t)(P_{11} B_{11} + \sum_{j=2}^{N} P_{1j}(1-B_{1j})) +\\
    &\qquad \sum_{i=2}^{N} z_i(t)\left(P_{i1}B_{i1} + \frac{P_{ii}(1-B_{ii})}{N-1}\right)
  \end{align*}

  Since $z_i(t)$ are known constants and we are optimizing over all
  $P_{ij}$ such that $\sum_j P_{ij} = 1$, we observe that each term in
  brackets that is multiplying $z_i(t)$ is a convex combination of
  terms containing $B_{ij}$. To maximize a convex combination, we set
  the coefficient of the largest term to be 1. This gives us the
  stated result.
\qed

\subsection{Simulation details}\label{sec:simDetails}

All of the regret and population proportion curves generated for comparison of ETC and TS policies in the context of an unknown rewards matrix have been done after averaging over 1000 simulations and letting each of the simulations run for 1000 time steps.

In each simulation, the type of the arriving user is sampled from a Bernoulli distribution with the probability of type 1 user arriving being $z_1(t)$ and the user being type 2 otherwise. The arm is then chosen according to the prescribed policy (ETC, TS or optimal). The arms and type together specify an element of the rewards matrix. The reward is thus sampled from a Bernoulli RV with this element as the mean. The population of users is updated according to the CID or DID model (as specified in Section \ref{sec:model}) and then we move on to the next time step where the above sequence of events is repeated.

%\section*{Acknowledgment}

\end{document}